\documentclass[10pt,twocolumn,letterpaper]{article}

\usepackage{wacv}
\usepackage{times}
\usepackage{epsfig}
\usepackage{graphicx}
\usepackage{amsmath}
\usepackage{amssymb}
\usepackage{booktabs}
\usepackage{cite, multirow}
\usepackage{multirow,epsfig,pdflscape,graphicx,amsmath,amssymb}
\usepackage{float}
\usepackage{amsmath,amssymb,amsfonts}
\usepackage{textcomp}
\usepackage{mathrsfs}
\usepackage{empheq}
\usepackage{rotating}
	
\usepackage{color, colortbl}
\definecolor{LightCyan}{rgb}{0.88,1,1}
\usepackage[table]{xcolor}
\usepackage{url}            % simple URL typesetting
\usepackage{booktabs}       % professional-quality tables
\usepackage{nicefrac}       % compact symbols for 1/2, etc.
\usepackage{microtype}      % microtypography
\usepackage{lipsum}
\usepackage{cite,multirow}
\usepackage{array, caption, tabularx,  ragged2e,  booktabs}
\usepackage{longtable}
\usepackage{textcomp}
\usepackage{siunitx}
\usepackage{array}
\usepackage{algorithm,algpseudocode}
\usepackage{xcolor,comment}

\usepackage{times}
\usepackage{epsfig}
\usepackage{amsmath}
\usepackage{amssymb}
\usepackage{amsthm}
\usepackage{bm}
\usepackage{appendix}
\usepackage{etoolbox}
\usepackage{soul}
\usepackage{tabularx}
\usepackage{caption}
\usepackage{subcaption,xspace}
\usepackage[breaklinks,colorlinks,bookmarks=false]{hyperref} %removed pagebackref
\usepackage[capitalize]{cleveref}

\theoremstyle{definition}

\usepackage[inline]{enumitem}

\usepackage[capitalize]{cleveref}
\crefname{section}{Sec.}{Secs.}
\Crefname{section}{Section}{Sections}
\Crefname{table}{Table}{Tables}
\crefname{table}{Tab.}{Tabs.}

\BeforeBeginEnvironment{appendices}{\clearpage}

\newcolumntype{Y}{>{\centering\arraybackslash}X}

\def\ramya#1{{\color{blue}{\bf [Ramya:} {\it{#1}}{\bf ]}}}

\usepackage{graphicx} 
\usepackage{caption}

\def\etal{{et al.\xspace}}
\def\ood{{OOD\xspace}}

\newcolumntype{M}[1]{>{\centering\arraybackslash}m{#1}}
\newcolumntype{N}{@{}m{0pt}@{}@{}}

\usepackage{enumitem}

\newcommand{\myfirstpara}[1]{\noindent \textbf{#1:}}
\newcommand{\mypara}[1]{\vspace{0.2em} \myfirstpara{#1}}

\graphicspath{{images/}}
\graphicspath{{./imgs/}}

\def\nll{\textsc{nll}\xspace}
\def\cnn{\textsc{cnn}\xspace}
\def\dnn{\textsc{dnns}\xspace}

\def\dnns{\textsc{dnn}s\xspace}
\def\sota{\textsc{sota}\xspace}

\def\mdca{\textsc{mdca}\xspace}
\def\sce{\textsc{sce}\xspace}
\def\ece{\textsc{ece}\xspace}

\def\ts{\textsc{ts}\xspace}
\def\ood{\textsc{ood}\xspace}
\def\dca{\textsc{dca}\xspace}
\def\mmce{\textsc{mmce}\xspace}
\def\dc{\textsc{dc}\xspace}
\def\ls{\textsc{ls}\xspace}

\def\flsd{\textsc{flsd}\xspace}
\def\te{\textsc{te}\xspace}
\def\ema{\textsc{ema}\xspace}

\makeatletter
\renewcommand\section{\@startsection{section}{1}{\z@}%
                       {-8\p@ \@plus -4\p@ \@minus -4\p@}%
                       {6\p@ \@plus 4\p@ \@minus 4\p@}%
                       {\normalfont\large\bfseries\boldmath
                        \rightskip=\z@ \@plus 8em\pretolerance=10000 }}
\renewcommand\subsection{\@startsection{subsection}{2}{\z@}%
                       {-8\p@ \@plus -4\p@ \@minus -4\p@}%
                       {6\p@ \@plus 4\p@ \@minus 4\p@}%
                       {\normalfont\normalsize\bfseries\boldmath
                        \rightskip=\z@ \@plus 8em\pretolerance=10000 }}
\renewcommand\subsubsection{\@startsection{subsubsection}{3}{\z@}%
                       {-4\p@ \@plus -4\p@ \@minus -4\p@}%
                       {-1.5em \@plus -0.22em \@minus -0.1em}%
                       {\normalfont\normalsize\bfseries\boldmath}}
\makeatother
\floatname{algorithm}{Procedure} 
\usepackage[accsupp]{axessibility}

\def\wacvPaperID{1560} % Enter the WACV Paper ID here

\wacvalgorithmstrack   % Uncomment this line if you are submitting to the Algorithms Track.
\wacvfinalcopy % *** Uncomment this line for the final submission

\ifwacvfinal
\begin{document}

%%%%%%%%% TITLE
\title{Calibrating Deep Neural Networks using Explicit Regularisation and Dynamic Data Pruning}
%\title{Regularisation and Dynamic Data Pruning can calibrate DNNs better.}
%\title{An Efficient Deep Neural Network calibration through explicit regularization and dynamic data pruning}

\author{Rishabh Patra$^{1}$\textsuperscript{\textsection} \quad  Ramya Hebbalaguppe$^{2}$\textsuperscript{\textsection} \quad  Tirtharaj Dash$^{1}$ \quad Gautam Shroff$^{2}$ \quad Lovekesh Vig$^{2}$\\
$^{1}$ APPCAIR, BITS Pilani, Goa Campus \quad
$^{2}$ TCS Research, New Delhi\\
%$^{*}$ First author\\
% {\tt\small f20180348@goa.bits-pilani.ac.in, ramya.hebbalaguppe@tcs.com}
% {\tt\small f20180348@goa.bits-pilani.ac.in}
% {\tt\small ramya.hebbalaguppe@tcs.com}
% {\tt\small tirtharaj@goa.bits-pilani.ac.in}
% {\tt\small gautam.shroff@tcs.com}
% {\tt\small lovekesh.vig@tcs.com}
}
\maketitle 
\begingroup\renewcommand\thefootnote{\textsection}
\footnotetext{Equal contribution}
\endgroup

%\endgroup

\begin{abstract}
  Deep neural networks (\dnn) are prone to miscalibrated predictions, often exhibiting a mismatch between the predicted output and the associated confidence scores. Contemporary model calibration techniques mitigate the problem of 
    overconfident predictions by pushing down the confidence of the winning class while increasing the confidence of the remaining classes across all test samples. However, from a deployment perspective an ideal model is desired to (i) generate well calibrated predictions for high-confidence samples with predicted probability say $>0.95$ and 
    (ii) generate a higher proportion of legitimate high-confidence samples. To this end, we propose a novel regularization technique that can be used with classification losses, leading to state-of-the-art calibrated predictions at test time; From a deployment standpoint in safety critical applications, only high-confidence samples from a well-calibrated model are of interest, as the remaining samples have to undergo manual inspection. Predictive confidence reduction of these potentially ``high-confidence samples'' is a downside of existing calibration approaches. We mitigate this via proposing a \textbf{dynamic train-time data pruning} strategy which prunes low confidence samples every few epochs,  providing an increase in \textbf{confident yet calibrated samples}.  We demonstrate state-of-the-art calibration performance across image classification benchmarks, reducing training time without much compromise in accuracy. We provide insights into why our dynamic pruning strategy that prunes low confidence training samples leads to an increase in high-confidence samples at test time.
    
    % \footnote{Our confidence calibration looks at lighter training regime, as a followup of dark side of calibration is low refinement as mentioned by Singh \etal \cite{singh2021deep}, a measure of trust in classification. Lighter side implies training a model with fewer ``relevant" samples}
\end{abstract}

\section{Introduction}
\label{sec:intro}

A Deep Neural Network classifier  outputs a class probability distribution that represents the relative likelihoods of the data instance belonging 
to a predefined set of class-labels. Recent studies have revealed that these networks are often miscalibrated or that the predicted confidence does not align to the probability of correctness
~\cite{guo2017calibration,pereyra2017regularizing,Dirichlet}. This runs against the recent emphasis around trustworthiness of AI applications based on \dnns, and begs the question ``\emph{When can we trust \dnns predictions?}''
This question is pertinent given the potential applications where \dnns are a part of decision making pipelines.
% both in safety critical use-cases such as disease diagnosis in healthcare as also in effort-reduction use-cases such as automated segmentation of radiology images. 

It is important to note that only high-confidence, meaning high predicted probability events matter for real-world use-cases. For example, consider an effort-reduction
use-case where a DNN model is used to automatically annotate radiology images in bulk before these are passed on to doctors; high-confidence outputs are passed on directly, and others are sent for manual annotation; ideally, only a few of the latter
else the effort-reduction goal is not met. Alternatively, consider
a disease diagnosis use-case where a deep model is used to decide whether
to send a patient to a COVID ward or a regular ward (e.g. for tuberculosis etc.) based on an X-ray pending a more conclusive test \cite{mahajan2020coviddiagnosis}. Clearly, only high-confidence negative predictions should be routed to a regular ward; maximising such
high-confidence, calibrated predictions is important; else, the purpose
of employing the deep model, i.e., reducing the load on the COVID ward, is not achieved. In both cases, low-confidence predictions cannot influence decision-making,
as such samples will be routed to a human anyway. Such applications demand a trustworthy, well-calibrated AI model, as overconfident and incorrect predictions can either prove fatal or deviate from the goal of saving effort. Additionally, the focus has to be on high-confidence samples, both in
their calibration as well as increasing their frequency. Efforts to
calibrate \textit{all} samples may not always align with this goal.

\mypara{Pitfalls of modern DNN confidence outcomes} 
Guo \emph{et. al}\cite{guo2017calibration} make an observation 
that poor calibration of neural networks is linked to overfitting of 
the negative log-likelihood (\nll) during training. 
The \nll loss in a standard supervised classification  
for an instance-label pair $(\mathbf{x}', y')$ sampled from a data distribution $\mathcal{P}_{data}$, is given as:
$\mathcal{L}_{CE} = - \log \hat{p}(y = y' | \mathbf{x}')$
% $\mathcal{L}_{CE} = -\log \hat{p}({y_i}|\mathbf{x}_i)$.
The \nll loss is minimized when for each $\mathbf{x}'$, 
$\hat{p}({y = y'}|\mathbf{x}') = 1$, 
whereas the classification error is minimized when 
$\hat{p}({y = y'}|\mathbf{x}') > \hat{p}({y \neq y'}|\mathbf{x}')$. 
Hence, the \nll can be positive even when the classification error is $0$, which causes models trained with $\mathcal{L}_{CE}$ to overfit to the \nll objective, leading to overconfident predictions and poorly calibrated models.
Recent parametric approaches involve temperature-scaling \cite{guo2017calibration}, which scale the pre-softmax layer logits of a trained model in order to reduce confidence. Train-time calibration methods include: \mmce \cite{kumarpaper}, label-smoothing (\ls) \cite{labelsmoothinghelp}, Focal-loss  \cite{focallosspaper}, \mdca \cite{StitchInTime} which add explicit regularizers to calibrate models during training. While these methods perform well in terms of reducing the overall Expected calibration Error (\ece)\cite{guo2017calibration} and scaling back overconfident predictions, they have two undesirable consequences: 
(1) From a deployment standpoint, only high confidence samples are of interest, as the remaining samples undergo manual inspection. Reducing model confidence in turn reduces the number of such high confidence samples, translating into more human effort; 
(2) Reduction of confidences compromises the separability of correct and incorrect predictions\cite{singh2021deep}. % , i.e., it harms the refinement of the model's predictions 

Further, train-time calibration methods require retraining of the models for recalibration. With the recent trend in  increasing overparametrization of models and training on large corpus of data, often results in high training times, reducing their effectiveness in practical settings. In this work, we investigate an efficient calibration technique, which not only calibrates \dnn models, but does so in a fraction of the time as compared to other contemporary train-time calibration methods. Additionally, we explore a practical setting where instances being predicted with a confidence greater than a user-specified threshold (say $95\%$) are of interest, as the others are routed to a human for manual screening. In an effort to reduce manual effort, we focus on increasing the number of such high confidence instances, and focus on calibrating these high confidence instances effectively.

\mypara{Contributions} We make the following key contributions:
\begin{enumerate*}[label=\textbf{(\arabic*)}]%[leftmargin=*, topsep=0.2em, itemsep=-0.2em]
%\mypara{Mitigating Overconfidence effectively} We make the following key contributions to repair miscalibration, with the goal of increasing the fraction of calibrated, legitimately high-confident predictions by reducing training times through dynamic data pruning strategy across models:

    \item We introduce a differentiable loss term that enforces calibration by reducing the difference between the model's predicted confidence and it's accuracy.
    \item We propose a dynamic train-time pruning strategy that leads to calibrated predictions with reduced training times. Our proposition is to prune out samples based on their predicted confidences at train time, leading to a reduction in training time without compromising on accuracy. We provide insights towards why dynamic train-time pruning leads to legitimate high confidence samples and better calibrated models. %in a fraction of the training time, and we provide theoritical support for the same. As an example, our proposed pruning strategy on CIFAR-10 .......
    %\item % Hence, we coin our approach ``\textit{Uncompromising calibrator}". \textcolor{red}{Is this still applicable?}
    % \item The goal is also to obtain calibrated models with a reduced train-time, a practical training regime suited for a low-GPU setting, we aim to obtain calibrated models using a fraction of input samples from the dataset dynamically. For example, we propose a confidence pruning strategy to show roughly using 50\% of the samples on CIFAR100 dataset, we achieve a not only a well-calibrated model but also achieve this without trade-off in accuracy. \textcolor{red}{Furthermore, we reduce the train time by a factor ..... Rishabh to fill this}
    
\end{enumerate*}

%The remainder of the paper is organised as follows: Section \ref{sec:dnn_calibration} defines \dnn confidence calibration, the metrics used to measure the confidence calibration of a Neural Network model. In Section \ref{sec:proposed_approach} we provide the details of our auxiliary loss and dynamic pruning at training-time of \dnns. Section \ref{sec:empirical_evaluation} presents empirical evaluations and comparison with the current state-of-the-art (\sota). Finally, we discuss related calibration methods in Sec. \ref{sec:relWorks} and their connections to our proposed method and conclude with the key takeaways.

\section{Related Work}
\label{sec:relWorks}

The practical significance  of the calibration problems addressed in this paper has resulted in a  significant body of prior literature on the subject. Existing solutions  employ either train-time calibration or post-hoc calibration.  Prior attempts for train-time calibration entail training on the entire data set while our algorithm aims at pruning less important samples to achieve high frequency of \textit{confident yet calibrated samples}, cutting down both on training time and subsequently compute required to train a calibrated model. 

% (instrinsic calibration)  : is not same as train time caibration - FL is one because it implicitly increases entropy on L_CE
% (extrinsic calibration).

% Train-time calibration integrate model calibration during the training procedure while a post-hoc calibration method utilizes a hold-out set to tune the calibration measures.

% \smallskip
\mypara{Calibration on Data Diet} 
Calibrating  \dnns  builds trust in the model's predictions, making them more suitable for practical deployment. However, we observe that \dnns have been getting deeper and bulkier, leading to high training times. \cite{datadietpaper} make a key observation that not all training samples contribute equally to the generalization performance of the \dnn at test time.
Choosing a core-set of instances that adequately represent the data manifold directly translates to lower training times without loss in performance.  \cite{forgettingpaper} mark instances that are ``forgotten" frequently while training, subsequently identifying such samples to be influential for learning, hence forming the core-set for training.

% Core-set selection is appealing from a practical standpoint as it reduces the carbon footprint.
% Further \cite{forgettingpaper} use forgetting scores to mark train-time instances that are ``forgotten" frequently during training time, subsequently identifying a core-set of such important samples that are influential for learning. However, these works focus only on the test time accuracy of these models, and do not make any note of the calibration effects of choosing a core-set.

% Choosing a core-set of instances implies using fewer instances than is available in the trainset to train the \dnns. This directly implies lower training times, lower carbon footprint and attractive from a practical deployment. 
Inspired from \cite{datadietpaper}, we hypothesize that not all samples would contribute equally to calibrating the model as well. We however differ in our approach to identifying important samples by choosing a dynamic pruning strategy. Our strategy dictates that samples which have low predicted confidence over multiple training epochs hamper the calibration performance, and thus shall be pruned.

% \smallskip
\mypara{Train-Time Calibration}  
% The Brier score was one of the earliest train-time methods proposed for  calibrating binary probabilistic forecasts \cite{brierloss}.
 %Models trained with Negative-Log-Likelihood (\nll) have been shown to be susceptible to over-confident predictions  \cite{guo2017calibration} which necessitates re-calibration. 
 A popular solution  to mitigate overconfidence is to use additional loss terms with the \nll loss: this includes using an entropy based regularization term  \cite{pereyra2017regularizing} or Label Smoothing \cite{labelsmoothinghelp} (\ls) \cite{originallabelsmoothing} on soft-targets. Recently, implicit calibration of \dnns was demonstrated\cite{ogfocalloss} via focal loss\cite{focallosspaper} which was shown to reduce the KL-divergence between predicted and target distribution whilst increasing the entropy of the predicted distribution, thereby preventing overconfident predictions. An auxillary loss term for model calibration \dca was proposed by Liang \etal \cite{dcapaper} which penalizes the model when the cross-entropy loss is reduced without affecting the accuracy. \cite{kumarpaper} propose to use \mmce calibration computed using RHKS~\cite{rkhskernel}. % \cite{maronas2021calibration} evaluate MixUp data augmentation for calibrating DNNs only to find that Mixup does not necessarily improve calibration.

% \smallskip
\mypara{Post-Hoc Calibration} 
 Post-hoc calibration typically uses a hold-out set for calibration. Temperature scaling (\ts) \cite{platt1999probabilistic} that divides the model logits by a scaling factor to calibrate the resulting confidence scores. The downside of using \ts for calibration is reduction in confidence of every prediction \cite{can-u-trust}, including the correct ones. Dirichlet calibration (\dc) is derived from Dirichlet distributions and generalizes the Beta-calibration \cite{beta-cal-paper} method for binary classification to a multi-class one. Meta-calibration propose differentiable ECE-driven  calibration to obtain well-calibrated and highly-accuracy models \cite{bohdal2021meta}. %Islam \etal \cite{islamclass} propose class-distribution-aware \ts and \ls that can be used as a post-hoc calibration. They use a class-distribution aware vector for \ts/\ls to fix the overconfidence. \cite{local-TS} propose a spatially localized calibration approach targeting semantic segmentation.

%\mypara{Calibration Through \ood Detection} 

\begin{comment}
\cite{hein2019relu} show that one of the main reasons behind the overconfidence in \dnns is the usage of ReLu activation that gives high confidence predictions when the input sample is far away from the training data. They mitigate this via data augmentation using adversarial training, which enforces low confidence predictions far away from the training data. \cite{guo2017calibration} analyze the effect of a \dnn's width, depth, batch normalization, and weight decay on the calibration. \cite{ood-spectral} use spectral analysis on initial layers of a \cnn to determine \ood sample and calibrate the \dnn for image segmentation task. We refer the reader to \cite{hendrycks2018deep, devries2018learning, padhy2020revisiting, meronen2020stationary} for other representative works on calibrating a \dnn through \ood detection.

\end{comment}

%Intuitively, 

\section{A Preliminary on DNN calibration}
\label{sec:dnn_calibration}

Let $\mathcal{P}_{data}$ denote the probability distribution
of the data.
Each dataset (training or test) consists of $(\mathbf{x},y)$ pairs
where each $(\mathbf{x},y) \sim \mathcal{P}_{data}$ (i.i.d. assumption).
Let a data instance $\mathbf{x}$ be multi-dimensional, that is,
$\mathbf{x} \in \mathbb{R}^{h\times w\times d}$
and $y \in \mathcal{Y}$ where $\mathcal{Y}$ is a set of
$K$-categories or class-labels: $\{1,2,\dots,K\}$.
We use $\mathcal{N}$ to denote a trained neural network with
structure $\pi$ and a set of parameters $\bm{\theta}$.
It suffices to say that:
$\mathcal{N}$ takes a data instance $\mathbf{x}$
as input and outputs a conditional probability vector
representing the probability distribution over $K$ classes:
$\hat{\bm{y}} = [\hat{y}_1,\dots,\hat{y}_K]$,
% Change this to \hat{\bm{y}} = \hat{y}^{(1)}, \dots \hat{y}^{(k)}? And \hat{y}_i denotes the predicted class for an instance label pair \mathbf{x}_i, y_i
where $\hat{y}_k$ denotes the predicted probability that $\mathbf{x}$
belongs to class $k$, that is, $\hat{y}_k = \hat{p}(y=k|\mathbf{x})$. 
We write this as: $\mathcal{N}(\mathbf{x}; (\pi,\bm{\theta})) = \hat{\bm{y}}$.
Further, we define the predicted class-label for $\mathbf{x}$
as:
\[
    \hat{y} = \arg \max_{i} (\hat{\bm{y}}),
\] 
and the corresponding confidence of prediction as:
\[
    c = \max(\hat{\bm{y}}).
\]

% \begin{comment}
\subsection{Over-confident and Under-confident Models}
Ideally, the model's predicted vector of probabilities should represent
ground truth probabilities of the correctness of the model.
For example: If the model's predicted probability for a class $k \in \{1,\dots,K\}$ is $0.7$, then we would expect that given 100 such predictions
the model makes a correct prediction in 70 of them.
Cases where the number of correct predictions is greater than 70 imply
an underconfident model. Similarly, cases where the number of correct
predictions are less than 70 imply an overconfident model.
Mathematically, for a given instance-label pair $(\mathbf{x}, y) \sim \mathcal{P}_{data}$,
$P(y = k | \hat{y}_k = s_k) > s_k$ indicates an underconfidence, and similarly, $P(y = k | \hat{y}_k = s_k) < s_k$ indicates overconfidence.
Here $P(y=k | \hat{y}_k = s_k)$ implies the probability of the model correctly predicting that the instance belongs to class $k$, given that its predicted probability for the instance belonging to class $k$ is $s_k$ (probability vector $\textbf{s}$ outcome of softmax).
Overconfident models are certainly undesirable for real world use-cases.
It may be argued that underconfident models are desirable, as given
a predicted probability of class $k$ as $s_k$, we infer that the
probability of correctness is $\geq s_k$.
We argue that while underconfident models are certainly more desirable
than overconfident models, it may be unappealing in certain use-cases.
Consider the COVID-19 use case again. If we decide a threshold of $0.95$
of negative prediction, ie, if the model classifies an instance as
negative with a probability $\leq 0.95$, the instance undergoes manual
annotation. Now consider an instance which has been classified negative with a probability of $0.51$. Underconfidence of the model implies that the probability of correctness is $\geq 0.51$, but this statement gives
the doctors no information about the probability of correctness
crossing their pre-defined threshold, and hence, this instance shall
undergo manual inspection regardless of the true probability of correctness.
We argue that if the model had been \textit{confident}, then a
predicted probability of $0.51$ would have been inferred as the
probability of correctness, and thus, the instance would have undergone
manual screening, whereas predicted probability of $0.96$ would have
not undergone manual screening. It is likely that an underconfident model
would have predicted negative for the same instance with a probability
$\leq 0.95$, thus increasing the human effort required for annotation.

\subsection{Confidence Calibration}
\label{subsec:conf_calib}
We focus on confidence calibration for supervised
multi-class classification with deep neural networks.
We adopt the following from Guo \textit{et al.}~\cite{guo2017calibration}: \textbf{Confidence calibration}
A neural Network $\mathcal{N}$ is confidence calibrated
if for any input instance, denoted by  $(\mathbf{x},y)$ and 
for any $\omega \in [0, 1]$:
\begin{equation}\label{eq:conf_calib}
    p(\hat{y} = y ~|~ c = \omega) = \omega
\end{equation}
% \end{definition}
% It is to be noted that stronger definitions of calibration exist such as Multiclass Calibration, Classwise Calibration \cite{Dirichlet}.
% However, throughout this paper, our focus remains on confidence calibration, as shown in Eq.~\eqref{eq:conf_calib}. 
Intuitively, the definition above means that
we would like the confidence estimate of the prediction to be representative
of the true probability of the prediction being accurate.
%For example, given 100 predictions from the deep model, each with an associated confidence of $0.8$, we expect that 80 predictions should be accurate. 
However, since $c$ is a continuous random variable, computing the probability is
not feasible with finitely many samples. To approximate this, 
a well-known empirical metric exists, called the \textit{expected calibration error}~\cite{guo2017calibration}, which measures the miscalibration
of a model using the expectation of the difference between its 
confidence and accuracy:
\begin{equation}
    \ece = \mathbb{E}_c[~|~ p(\hat{y} = y | c = \omega) - \omega ~|~].
   % \ref{eq:conf_calib}
\end{equation}
In implementations, the above quantity is approximated by
partitioning predictions of the model into $M$-equispaced bins,
and taking the weighted average of the bins' accuracy-confidence
difference. That is,
\begin{equation}
\label{eq:ece}
    \ece = \sum_{m=1}^M \frac{|B_m|}{n} |a_{m} - c_{m}|,
\end{equation}
where $n$ is the number of samples. The gap between the accuracy and the
confidence per bin represents the \textit{calibration gap}.
Calibration then can be thought of as an optimisation problem, where model parameters $\bm{\theta}$ are modified to obtain the minimum gap. That is,
\begin{equation}
    \bm{\theta}' = \arg \min_{\bm{\theta}} \ece.
\end{equation}

% Since we have only a finite number of samples, the \ece cannot in practice be computed using this definition. Instead, the interval $[0, 1]$ is divided into $M$ equispaced bins, where the $i^{th}$ bin is the interval $\left(\frac{i-1}{M}, \frac{i}{M}\right]$. Let $B_i$ denote the set of samples belonging to the $i^{th}$ bin. The accuracy of this bin is calculated as $A_i = \frac{1}{|B_i|}\sum_{(\mathbf{x}, y) \in B_i} \mathbb{I} (y = \hat{y})$, similarly, the confidence of the bin is calculated as $C_i = \frac{1}{B_i} \sum_{(\mathbf{x}, y) \in B_i} \max \hat{\bm{y}}$. The \ece can then be approximated as a weighted average of the absolute difference between the accuracy and confidence of each bin: $\ece = \sum_{i=1}^M \frac{1}{|B_i|} |A_i - C_i|$. 
Evaluations of \dnn calibration via \ece suffer from the following shortcomings: (a) \ece does not measure the calibration of all the classes probabilities in the predicted vector; 
(b) the \ece score is not differentiable due to binning and cannot be directly optimized;
(c) the \ece metric gives equal importance to all the predictions despite their confidences. 
However, in a practical setting, the instances predicted with a high-confidence are of interest, and thus, calibration of high-confidence instances should be given relatively more importance. 

\begin{comment}

\mypara{Static Calibration Error (\sce)} \cite{StitchInTime, nixon2019measuring} look at the classwise extension of the \ece metric, and is calculated as:
\begin{equation}
    \sce = \frac{1}{K} \sum_{i=1}^{M} \sum_{j=1}^{K} \frac{1}{|B_{i, j}|} |A_{i, j} - C_{i, j}|
\end{equation}
where $K$ denotes the number of classes, $B_{i, j}$ denotes the number of samples in the $i^{th}$ bin belonging to the $j^{th}$ class. Further, $A_{i,j} = \frac{1}{B_{i,j}} \sum_{(\mathbf{x}, y) \in B_{i,j}} \mathbb{I} (\hat{y} = j)$ is the accuracy for the samples of $j^\text{th}$ class in the $i^{th}$ bin, and $C_{i,j} = \frac{1}{B_{i,j}} \sum_{(\mathbf{x}, y) \in B_{i,j}} \hat{p}(y = j | \mathbf{x})$ or average predicted probability for the $j^\text{th}$ class in the $i^\text{th}$ bin.

\end{comment}
\section{Proposed Approach}
\label{sec:proposed_approach}

Our goal is to remedy both: the overconfident and underconfident decisions from a \dnn based classifier while reducing the training time required to obtain a well-calibrated model.
We achieve this by employing a combination of implicit and explicit regularisers. First, we implicitly regularise the model via a focal loss \cite{focallosspaper}; Second, we propose an auxiliary Huber loss to penalise for calibration error between the avg. confidence and avg. accuracy of samples in a mini-batch. Our total loss is the weighted sum of the classification loss with implicit regularization and an explicit regulariser as discussed in Sec \ref{subsec:explicitCalib}. Third, we propose a dynamic pruning strategy in which low confidence samples from the training set are removed to improve the training regime and enforce calibration of the highly confident predictions (explained in \cref{subsec:DynaPrune}) that details the procedure. \cref{proc:learn} provides an outline of our proposed confidence calibration with dynamic data pruning detailed in \cref{proc:emaprune}.
% Please refer to the dynamic train-time pruning strategy
\subsection{Improving calibration via implicit regularisation}
\label{subsec:implicitCalib}
Minimising the focal loss \cite{focallosspaper, ogfocalloss} induces an effect of adding a maximum-entropy regularizer, which prevents predictions from becoming overconfident.
Considering $\bm{q}$ to be the one-hot target distribution for the instance-label pair $(\mathbf{x}, y)$, and $\hat{\bm{y}}$ to be the predicted distribution for the same instance-label pair, the cross-entropy objective forms an upper bound on the KL-divergence between the target distribution and the predicted distribution. That is, $\mathcal{L}_{CE} \geq \text{KL}(\bm{q} || \hat{\bm{y}})$. 
The general form of focal loss can be shown to form an upper bound on the KL divergence, regularised by the negative entropy of the predicted distribution $\hat{\bm{y}}$, $\gamma$ being the regularisation parameter \cite{mukhoti2020calibrating}. 
Mathematically, $\mathcal{L}_{FL} \geq \text{KL}(\bm{q} || \hat{\bm{y}}) - \gamma \mathbb{H}[\hat{\bm{y}}])$.
Thus focal loss provides implicit entropy regularisation to the neural network reducing overconfidence in \dnns.
\begin{equation}
\label{eq:fl}
\mathcal{L}_{FL}(\mathbf{x}, y) = (1-\hat{p}(y = y | \mathbf{x}))^\gamma \log \hat{p}(y=y | \mathbf{x}).
\end{equation} 
$\mathcal{L}_{FL}$ is the focal loss on an instance-label pair $(\mathbf{x},y)$.
%However, we investigate the calibration error with inclusion of explicit regulariser as described in the \cref{subsec:explicitCalib}

\subsection{Improving calibration via explicit regularisation}
\label{subsec:explicitCalib}
We aim to further calibrate the predictions of the model via explicit regularization as using $\mathcal{L}_{FL}$ alone can result in underconfident \dnns. \dca \cite{dcapaper} is an auxiliary loss that can be used alongside other common classification loss terms to calibrate DNNs effectively. However, the use of the $\mathcal{L}_1$ term in \dca renders it non-differentiable, and sometimes, fails to converge to a minima, thereby hurting the accuracy of the model's predictions. Unlike \dca \cite{dcapaper}, we propose using Huber Loss \cite{huber1992robust} to reap its advantages over $\mathcal{L}_1$ and $\mathcal{L}_2$ losses, specifically its differentiability for gradient based optimization. It follows that the latest gradient descent techniques can be applied to optimize the combined loss, leading to better minimization of the difference between the model's confidence and accuracy, and subsequently, improved calibration of the model's predictions. An added benefit is the reduced sensitivity to outliers that is commonly observed in $\mathcal{L}_2$ losses. Mathematically, the Huber loss is defined as:
\begin{equation}
    \mathcal{H}_{\alpha}(x) =
    \begin{cases}
    \frac{1}{2}x^2, &\text{for } |x| \leq \alpha \\
    \alpha (|x - \frac{1}{2}\alpha|), &\text{otherwise}
    \end{cases}
\end{equation}

where $\alpha \in [0, 1]$ is a hyperparameter to be tuned that controls the transition from $\mathcal{L}_1$ to $\mathcal{L}_2$.

Specifically, for a sampled minibatch, $B = \{(\mathbf{x}_j, y_j)\}_{j=1}^{|B|}$, %\subseteq \mathcal{D}$, 
the proposed auxiliary loss term is calculated as,
\begin{equation}
\label{eq:huber}
    \mathcal{L}_H = \mathcal{H}_{\alpha}\bigg( \frac{1}{|B|}\sum_{i=1}^{|B|} c_i - \frac{1}{|B|}\sum_{i=1}^{|B|} \mathbb{I}(\hat{y}_i = y_i)  \bigg).
\end{equation}

Here, $\alpha \in \mathbb{R}^+$ is a hyperparameter that controls the transition from $\mathcal{L}_1$ to $\mathcal{L}_2$ loss. The optimal value of $\alpha$ is chosen to be the one that gives us the least calibration error. 
Thus, we propose to use a total loss $\mathcal{L}_{total}$ as the weighted summation of focal loss, $\mathcal{L}_{FL}$ and a differentiable, Huber loss, $\mathcal{L}_{H}$ which penalises difference in confidence and accuracy in a mini-batch. $\mathcal{L}_{total}$ takes the form:
\begin{equation}
\label{eq:fL+h}
    \mathcal{L}_{total} = \mathcal{L}_{FL} + \lambda \mathcal{L}_{H}.
\end{equation}
%where $\mathcal{L}_{FL}(\mathbf{x}_i, y_i) = \frac{1}{N}\sum_{i=1}^N(1-\hat{p}(y = y_i | \mathbf{x}_i))^\gamma log \hat{p}(y=y_i | \mathbf{x}_i)$ is the focal loss calculated on an instance-label pair $(\mathbf{x}_i, y_i)$ and $\mathcal{L}_{H}$ for a minibatch of instances is calculated as defined in Equation \ref{eq:huber}.

\subsection{Boosting confidences via train-time data pruning}
\label{subsec:DynaPrune}

Entropy regularization by using $\mathcal{L}_{FL}$ reduces the aggregate calibration error, but in the process, needlessly clamps down the legitimate high confidence predictions. This is undesirable from a deployment standpoint, where highly confident and calibrated predictions are what deliver value for process automation. The low confidence instances, i.e, instances where the model's predicted confidence is less than a  specified threshold can be routed to a domain expert for manual screening.
\cite{singh2021deep} highlight the clamping down on predictions across a wide range of datasets and different model architectures. Our experiment confirms this trend, showing a reduction of confidences globally across all classes. % of our experiments.
% \ramya{Rishabh, are you sure Singh discusses this, I thgt they just discuss refinement}

To mitigate this problem and make our model \emph{confident yet calibrated}, we propose a simple yet effective train-time data pruning strategy based on the \dnn's predicted confidences. 
Further, we modify the \ece metric to measure calibration of these highly confident samples. Specifically, for an instance-label pair in the test set: $(\mathbf{x}, y) \in D_{te}$ and a confidence threshold $\delta$, for any $\omega \in [0, 1]$:
\begin{equation}
    \ece(S_{\delta}) = \mathbb{E}_c\big[~|~ p(\hat{y} = y | c = \omega; \mathbf{x} \in S_{\delta})) - \omega ~|~\big]
\end{equation}

where $S_{\delta}$ is the set of test set instances that are predicted with a confidence $\geq \delta$ (defined formally in Def. \ref{def:highConfset}). Note that the pruning is performed after the application of the Huber loss regularizer, $\mathcal{L}_H$.

Our proposition to divide the training run into checkpoints, and at each checkpoint, the model's performance is measured over the training dataset. Given this observation and information retained over previous checkpoints, we make a decision on which instance-label pairs to use for training until the next checkpoint, and which instance-label pairs to be pruned out.
We preferentially prune out instances for which the model's current prediction is least confident.
Such a pruning strategy evolves from our hypothesis that as the training progresses, the model should grow more confident of its predictions. However, the growth of predicted confidence is uneven across all the train set instances, implying that the model remains relatively under-confident about its predictions over certain samples, and relatively overconfident on other samples.
Our hypothesis is that these relatively under-confident instances affect the model's ability to produce calibrated and confident results, thus justifying our pruning strategy. 
Since model confidences can be noisy, we propose to prune based on an Exponential Moving Average (\ema) score, as opposed to merely the confidences, to track evolution of the confidence over multiple training epochs.
The \ema-score for a data instance $\mathbf{x}_i$ at
some $j^{\mathrm{th}}$-epoch during training, denoted by $e^{(j)}_i$, is calculated
using the moving average formula:
\begin{equation}
\label{eq:emascore}
    e^{(j)}_i = \kappa c^{(j)} + (1-\kappa) e^{(j-1)}_i
\end{equation}
where $\kappa \in [0,1]$
and $c^{(j)}$ is the confidence of prediction for the
data instance $\mathbf{x}_i$ at epoch $j$.
Before the model training starts, i.e. at $j = 0$, 
$e^{(j)}_i = 0$.
At any epoch $j$, we associate each data instance in a dataset
with its corresponding \ema-score as: 
$(\mathbf{x}_i,y_i,e^{(j)}_i)$.
We denote the set of data-instances with their \ema-scores
as $De = \{(\mathbf{x}_i,y_i,e^{(j)}_i)\}_{i=1}^{n}$,
where $n$ is the number of data-instances.
Procedure~\ref{proc:emaprune} provides pseudo-code for our proposed dynamic pruning method based on \ema-scores and the parameters of a
\dnn is learned using Procedure~\ref{proc:learn}.
In what follows, we provide some theoretical insights on our proposed technique.

\begin{algorithm}[!htb]
	\caption{\ema-based Low Confidence Pruning Procedure. The procedure takes as inputs: 
	a dataset of instances with their \ema scores, denoted by $De = \{(\mathbf{x}_i, y_i, e_i)\}_{i=1}^{n}$, and prune fraction parameter $\epsilon \in (0,100)$;
	and returns: a pruned dataset}
	\label{proc:emaprune}
	\begin{algorithmic}[1]
		\Procedure{PruneUsingEMA}{$De$, $\epsilon$}
		\For{every class $k$ in $1,\dots, K$}
            \State Let $De_{k}$ contain only the $k^{\mathrm{th}}$-class instances from $De$;
            \State $De = De \setminus De_{k}$
            \State Sort $De_{k}$ in ascending order by \ema scores;
            \State Let $De_{k,\epsilon}$ be the top-$\epsilon$ percentage of instances from $De_{k}$;
            \State Prune $De_{k}$ as: $De_{k} = De_{k} \setminus De_{k,\epsilon}$;
            \State $De = De \cup De_{k}$
        \EndFor
        \State \Return $De$
		\EndProcedure
	\end{algorithmic}
\end{algorithm}

\begin{algorithm}[!htb]
	\caption{Our pruning-based learning procedure. 
	The procedure takes as inputs: 
	A dataset of $n$ instances: $D = \{(\mathbf{x}_i, y_i)\}_{i=1}^{n}$,
	An untrained neural network $\mathcal{N}$ with structure $\pi$ and parameters $\bm{\theta}$,
	Maximum number of training epochs: $MaxEpochs$,
	Regularization parameter: $\lambda$;
	and returns: a trained model. An elaborate version of this procedure
	is in supplementary material.}
	\label{proc:learn}
	\begin{algorithmic}[1]
		\Procedure{TrainDNN}{$D$,$\mathcal{N}$,$\pi$,$\bm{\theta}$,$MaxEpochs$,$\lambda$,$\epsilon$,$\kappa$}
		
		\State Let $De = \{(\mathbf{x}_i, y_i, 0)\}_{i=1}^{n}$ 
        \State Initialise $\bm{\theta}$ to small random numbers
        \For{training epoch $ep$ in $\{1,\dots,MaxEpochs\}$}
            \ForAll minibatch $B_i \subset De$
                \State Compute average focal loss, $\mathcal{L}_{FL}$ 
                    \Comment{Eq.~\eqref{eq:fl}}
                \State Calculate Huber loss, $\mathcal{L}_{H}$ 
                    \Comment{Eq.~\eqref{eq:huber}}
                \State Calculate total loss, $\mathcal{L}_{total}$ 
                    \Comment{Eq.~\eqref{eq:fL+h}}
                \State Update $\bm{\theta}$ by minimising $\mathcal{L}_{total}$
            \EndFor 
            \State Update $De$ with updated \ema-scores
                \Comment{Proc.~\ref{proc:emaprune}}%, Eq.~\eqref{eq:emascore}}
        \EndFor
		\EndProcedure
	\end{algorithmic}
\end{algorithm}

Let $D_{tr} = \{(\mathbf{x}_i,y_i)\}_{i=1}^{m}$ and $D_{te} = \{(\mathbf{x}_i,y_i)\}_{i=1}^{n}$ represent the training and the test datasets, respectively, where each $(\mathbf{x}_i, y_i) \sim \mathcal{P}_{data}$ with i.i.d.
The confidence threshold parameter $\delta$ in the definition below
is empirical: It is primarily problem-specific and user-defined.
We further make the following observation.

\textbf{High-confidence instance: }\label{def:highConfset}
Given a trained neural network $\mathcal{N}$ with structure and
parameters $\pi$ and $\bm{\theta}$, respectively, 
and a threshold parameter $\delta \in (0,1]$,
a data instance $\mathbf{x}$ is called a high-confidence
instance if the confidence of prediction $c \geq \delta$
where $c = \max(\hat{\bm{y}})$ and 
$\hat{\bm{y}} = \mathcal{N}(\mathbf{x};(\pi,\bm{\theta}))$.
We denote a set of such high-confidence instances by a
set $S_{\delta}$.
% \end{definition}

% \begin{definition}[Set of High-confidence Samples]
% \label{def:set_of_high_conf_instances}
% We construct a set of High confidence instances from the test set w.r.t model parameters $\bm{\theta}$ and $\delta$; $S_{\delta} = \{\mathbf{x} | \mathbf{x} \in \mathcal{D}_{te} \wedge \mathbf{x} \text{ is a high confidence instance for parameters } \theta\}$ and $S_{p, \delta} = \{\mathbf{x} | \mathbf{x} \in \mathcal{D}_{te} \wedge \mathbf{x} \text{ is a high confidence instance for parameters } \theta_p\}$
% \end{definition}

% \begin{assumption}[1]
% \label{1}
% We consider $(\mathbf{x}, y) \in \mathcal{D}_{tr}$, where the predictive confidence, $c_{\theta}(\tilde{x}) = \argmin_{(\mathbf{x}, y) \in \mathcal{D}_{tr}} c_{\theta}(\mathbf{x})$ and also $\mathcal{D}_{te}$ follows the same probability distribution.
% \end{assumption}

%\begin{Observation}
Let $D_{tr}$ and $D_{te}$ be the training and test datasets.
Consider now two neural networks $\mathcal{N}$ and $\mathcal{N}'$
be two neural networks with the same structure $\pi$. In addition,
let's assume that the parameters of these two neural networks
are initialised to the same set of values before training
using an optimiser.
Let $\mathcal{N}$ be trained using $D_{tr}$ without pruning, resulting
in model parameters $\bm{\theta}$.
Let $\mathcal{N}'$ be trained using $D_{tr}$ with our proposed
train-time \textbf{dynamic} data pruning approach, resulting
in model parameters $\bm{\theta}'$.
Let $S_{\delta}$ and $S'_{\delta}$ 
denote the sets of high-confidence samples in
$D_{te}$ obtained using $\mathcal{N}$ and $\mathcal{N}'$,
respectively.
We claim that 
% if the predicted confidences of all x in D_tr is not the same, then
there exists a confidence-threshold $\delta \in [0,1]$
such that $|S'_{\delta}| \geq |S_{\delta}|$.
%\end{Observation}

The above claim implies that naively pruning out 
low confidence samples using 
our proposed pruning-based calibration technique leads to a relative
increase in the proportion of high-confidence samples during inference.
However, the reader should note that pruning during the initial stages
(initial iterations or epochs) of training could lead to the data
manifold to be inadequately represented by a neural network. 
One way to deal with this, as is done in our study, is to prune the
training set only after the network has been trained sufficiently, for example, when the loss function starts plateauing. It should be noted that the aim is to preserve relative class-wise imbalance after pruning. This is done to avoid skewing the train data distribution towards any class. We achieve such preservation by pruning the same fraction of instances across all classes.

\section{Empirical Evaluation}
\label{sec:empirical_evaluation}

%\section{Training methodology}
%\label{sec:tm}
\subsection{Datasets}
We detail the datasets used for training and evaluation in the supplementary material.
% We validate our proposed approach on benchmark datasets for image classification. We chose CIFAR-10/100 datasets
% \footnote{\url{https://www.cs.toronto.edu/~kriz/cifar.html}}, MendelyV2 (Medical image classification~\cite{kermany2018labeled}), SVHN\footnote{\url{http://ufldl.stanford.edu/housenumbers/}}, and Tiny ImageNet\footnote{\url{https://image-net.org/}}. 
% In all our experiments, we calibrate ResNet-50 \cite{resnetpaper} and measure the 
% calibration performances using our proposed calibration technique and several other existing techniques. For all experiments, the train set is split into $2$ mutually exclusive sets: (a) training set containing $90\%$ of samples and (b) validation set: $10\%$ of the samples. The same validation set is used for post-hoc calibration.

\subsection{Training Methodology}
\label{sec:tm}

For CIFAR10, we train the models for a total of $160$ epochs using an initial learning rate of $0.1$. The learning rate is reduced by a factor of $10$ at the $80^{th}$ and $120^{th}$ epochs. The \dnn was optimized using Stochastic Gradient Descent (SGD) with momentum $0.9$ and weight decay set at $0.0005$. Further, the images belonging to the trainset are augmented using random center cropping, and horizontal flips. For CIFAR100, the models are trained for a total of $200$ epochs with a learning rate of $0.1$, reduced by a factor of $10$ at the $100^{th}$ and $150^{th}$ epochs. The batch size is set to 1024. Other parameters for training on CIFAR100 are the same as used for CIFAR10 experiments. For Tiny-Imagenet, we follow the same training procedure used by \cite{focallosspaper}, with the exception that the batch size is set to 1024. For SVHN experiments, we follow the training methodology in \cite{StitchInTime}. For Mendeleyv2 experiments, we follow \cite{dcapaper} using a ResNet50 pre-trained on imagenet and finetuned on Mendeleyv2. The hyperparameters used were the same as used in \cite{dcapaper}.

We use higher batch sizes for Tiny-Imagenet and CIFAR100 than is common in practice. This is because of the classwise pruning steps in our proposed algorithms. With the example of CIFAR100, to prune out 10\% of the samples at every minibatch requires that at least 1000 samples be present in a minibatch, assuming random sampling across classes. Hence, we use a batch size of 1024 for CIFAR100. Similarly, for Tiny-Imagenet experiments, we prune out 20\% of the samples at every minibatch, requiring a batch-size of at least 1000. Hence we use a batch size of 1024 for Tiny-Imagenet experiments.

For the Huber loss hyperparameter $\alpha$, we perform a grid search over the values: $\{0.001, 0.005, 0.01, 0.05, 0.1\}$. The setting $\alpha=0.005$ gave the best calibration results across all datasets, and hence we use the same value for all the experiments. Other ablations regarding the Pruning Intervals, Regularization factor ($\lambda$), EMA factor ($\kappa$) can be found in section on Ablation Study.
All our experiments are performed on a single Nvidia-V100 GPU, with the exception of Tiny-Imagenet experiments, for which 4 Nvidia-V100 GPUs were used to fit the larger batch-size into GPU memory.

\subsection{Evaluation Metrics}

We evaluate all methods on standard calibration metrics: \ece along with the test error (\te). Recall that \ece measures top-label calibration error and \te is indicative of generalisation performance. A lower \ece and lower \te are preferred.  Additionally, to measure top label calibration, we report the \ece of the samples belonging to the $S_{95}$ and $S_{99}$ sets (Definition \ref{def:highConfset}). In \cref{tab:S95-all-methods}, we also report $|S_{95}|$ and $|S_{99}|$ as a percentage of the number of samples in the test set. Our method obtains near state-of-the-art top-label calibration without trading off accuracy/\te. We also visualise the calibration performance using reliability diagrams.

\subsection{Results}
\mypara{Calibration performance Comparison with \sota}  \cref{tab:sce-all-methods} compares calibration performance of our proposed methods against all the recent \sota methods. On the basis of the calibration metrics, we conclude that our proposed method improves  \ece score across all the datasets.  

\begin{table*}[!htb]
	\centering
	\resizebox{\linewidth}{!}{%
		\begin{tabular}{c||ccc|ccc|ccc|ccc|ccc|ccc||ccc}
			\toprule
			&  \multicolumn{3}{c|}{\textbf{BS \cite{brierloss}}} & \multicolumn{3}{c|}{\textbf{DCA \cite{dcapaper}}}   & \multicolumn{3}{c|}{\textbf{LS \cite{originallabelsmoothing}}} &
			\multicolumn{3}{c|}{\textbf{MMCE \cite{kumarpaper}}} &
			\multicolumn{3}{c|}{\textbf{FLSD \cite{focallosspaper}}}  &
			\multicolumn{3}{c||}{\textbf{FL + MDCA \cite{StitchInTime}}} &
			\multicolumn{3}{c}{\textbf{Ours ({FLSD+H+$\text{P}_{\text{EMA}}$})}} \\
			\multirow{-2}{*}{\textbf{Dataset}} &
			%\multirow{-2}{*}{\textbf{Model}} &

			ECE &
			TE &
			AUROC &
			
			ECE &
			TE &
			AUROC &
			
			ECE &
			TE &
			AUROC &
			
			ECE &
			TE &
			AUROC &
			
			ECE &
			TE &
			AUROC &
			
			ECE &
			TE &
			AUROC &
			
			ECE & 
			TE &    
			AUROC \\
			
			\hline
			\midrule
			CIFAR10  & 2.11 & 5.69 & 91.76 & 3.45 & 5.09 & 93.03 & 3.43 & 4.80 & 82.71 & 3.12 & \textbf{5.05} & 93.55 & 3.37 & 5.61 & 93.39 & 1.06 & 5.32 & \textbf{93.60} & \textbf{0.59} & 6.82 & 91.64       \\ \midrule
			CIFAR100 & 5.18 & 26.74 & 86.12 & 6.7 & 45.43 & 86.75 & 3.72 & \textbf{25.67} & 86.45 & 10.52 & 25.68 & \textbf{86.81} & 3.25 & 26.56 & 85.88 & 2.96 & 26.51 & 86.32 & \textbf{1.74} & 26.57 & 86.17   \\ \midrule
% 			\multirow{2}{*}{SVHN}     & ResNet20               &   &  &  &   &  &  &   &  &  &   &  &  &   &  &  &  &  &  &       \\
			SVHN & 1.22 & 3.36 & 88.50 & 0.71 & 3.83 & 92.04 & 4.25 & \textbf{3.25} & 84.57 & 1.86 & 3.46 & 87.75 & 12.58 & 3.61 & 88.38 & 7.96 & 3.74 & 89.84 & \textbf{0.38} & 3.99 & \textbf{90.79}     \\ \midrule
			Mendeley V2 & 22.99 & 27.04 & 64.91 & 21.42 & 22.59 & 56.60 & 15.81 & 23.39 & 66.19 & 21.73 & 25.48 & 72.19 & 12.59 & 31.41 & 71.99 & 17.09 & 27.08 & 73.48 & \textbf{11.69} & \textbf{22.27} & \textbf{75.96}   \\ \midrule
%			Kather5000                & ResNet34                &  32.76 & 8.64 & 16.50 &  27.13 & 4.49 & 16.80 &  47.35 & 16.63 & 17.90 &  57.87 & 22.03 & 19.30 &  41.39 & 13.45 & 16.60              \\ \midrule
			Tiny-ImageNet & 0.36 & 99.32 & 78.37 & 13.98 & 47.59 & \textbf{82.99} & 15.55 & \textbf{45.28} & 82.80 & 9.33 & 46.29 & 82.20 & 4.42 & 46.76 & 82.75 & 4.67 & 46.14 & 82.50 & \textbf{1.44} & 51.26 & 82.00         \\ \midrule
% 			20 Newsgroups             & Global-Pool CNN                &  725.82 & 13.71 & \textbf{25.93} &  719.83 & 15.30 & 28.07 &  731.31 & 12.69 & 28.63 &  940.70 & \textbf{4.52} & 30.80 &  \textbf{487.82} & 16.55 & 27.88         \\ 
			\bottomrule
		\end{tabular}%
	}
	\caption{Calibration measure \ece (\%) score), Test Error (\te) (\%) and AUROC (refinement) in comparison with various competing methods. We use $M=10$ bins for \ece calculation. We outperform most of the baselines across various popular benchmark datasets, and architectures in terms of calibration, while maintaining a similar accuracy and a similiar refinement (AUROC).}
	\label{tab:sce-all-methods}
%\vspace{-6mm}
\end{table*}

\begin{figure}[!htb]
    \centering
    \includegraphics[width=0.8\linewidth]{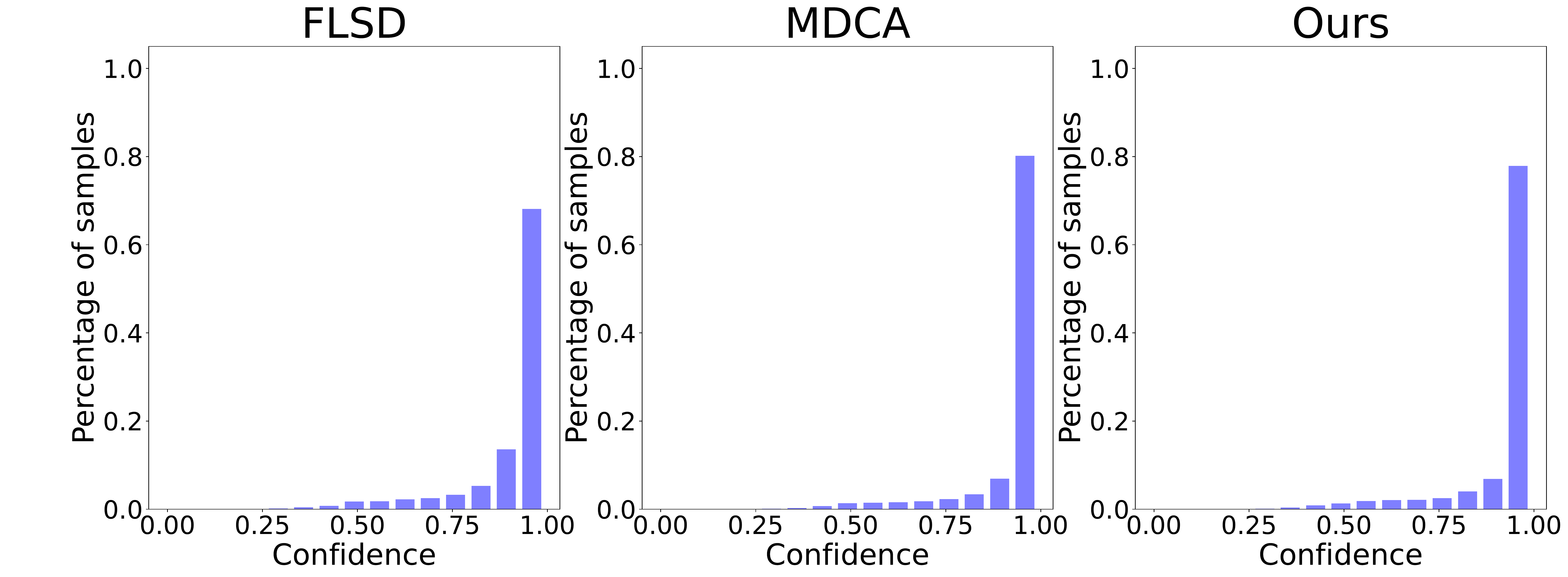}
    \caption{
       Histogram of confidences: Notice a rise in the number of instances falling in the last bin, ie, number of instances in the highest confidence bin in our method vs \flsd \cite{focallosspaper}. While the number of instances in the highest confidence bin is nearly equal in case of \mdca \cite{StitchInTime} and our method has better top-label calibration.
        }
        \label{fig:boostedHigh ConfSamples}
%\vspace{-5mm}
%\end{wrapfigure}
\end{figure}

\smallskip
\mypara{Top-label calibration (\ece)} For mission critical tasks, performance on those samples which are predicted with a high confidence are given higher importance than ones with low predicted confidence. \cref{tab:S95-all-methods} notes the \ece of the samples belonging to $S_{95}$ (this metric is named \ece($S_{95}$)). We also note the number of high confidence samples, $|S95|$ as a percentage of the total number of test samples. The \ece($S_{95})$ metrics show that our proposed method obtains near-perfect calibration on these high-confidence samples.  \cref{fig:boostedHigh ConfSamples} shows an increase in the number of instances falling in the highest confidence bin when moving from \flsd to our method. We also obtain better top-label calibration than \mdca\cite{StitchInTime} by $5\times$, despite the number of instances falling in the highest confidence bin is nearly equal.

\begin{table*}[!htb]
	\centering
	\resizebox{\linewidth}{!}{%
		\begin{tabular}{c|cc|cc|cc|cc|cc|cc||cc}
			\toprule
			&  \multicolumn{2}{c|}{\textbf{BS \cite{brierloss}}} & 
			\multicolumn{2}{c|}{\textbf{DCA \cite{dcapaper}}} & 
			\multicolumn{2}{c|}{\textbf{LS \cite{hinton2015distilling}}} &
			\multicolumn{2}{c|}{\textbf{MMCE \cite{kumarpaper}}} &
			\multicolumn{2}{c|}{\textbf{FLSD \cite{focallosspaper}}}  &
			\multicolumn{2}{c||}{\textbf{FL + MDCA \cite{StitchInTime}}} &
			\multicolumn{2}{c}{\textbf{Ours ({FLSD+H+$\text{P}_{\text{EMA}}$})}} \\
			\multirow{-2}{*}{\textbf{Dataset}} &

			ECE ($S_{95}$) &
			$|S_{95}|$ &
			
			ECE ($S_{95}$) &
			$|S_{95}|$ &
			
			ECE ($S_{95}$) &
			$|S_{95}|$ &
			
			ECE ($S_{95}$) &
			$|S_{95}|$ &
			
			ECE ($S_{95}$) &
			$|S_{95}|$ &
			
			ECE ($S_{95}$) &
			$|S_{95}|$ &
			
			ECE ($S_{95}$) &
			$|S_{95}|$  \\
			\hline
			\midrule
			CIFAR10 & 0.87 & 87.05 & 2.27 & \textbf{93.30}  & 3.01 & 84.82 & 1.89 & 92.83 & 2.06 & 60.27 & 0.80 & 76.71 & \textbf{0.12} & 74.94          \\ \midrule
			CIFAR100 & 2.92 & 45.24 & 1.5 & 15.13 & 1.29 & 28.7 & 5.44 & \textbf{58.96} & 1.22 & 20.12 & 1.35 & 22.23 & \textbf{0.01} & 33.1      \\ \midrule
% 			\multirow{2}{*}{SVHN}     & ResNet20               &   &  &  &   &  &  &   &  &  &   &  &  &  &         \\
			SVHN & 0.68 & 92.02 & 0.36 & 88.5 & 3.72 & 48.32 & 1.14 & \textbf{94.64} & 3.55 & 0.41 & 3.75 & 13.23 & \textbf{0.045} & 86.39        \\ \midrule
			Mendeley V2 & 21.95 & 74.84 & 22.04 & \textbf{93.10}  & 13.22 & 58.49 & 21.11 & 82.05 & 7.92 & 11.58 & 7.70 & 53.52 & \textbf{1.94} & 22.27    \\ \midrule
%			Kather5000                & ResNet34                &  32.76 & 8.64 & 16.50 &  27.13 & 4.49 & 16.80 &  47.35 & 16.63 & 17.90 &  57.87 & 22.03 & 19.30 &  41.39 & 13.45 & 16.60              \\ \midrule
			Tiny-ImageNet & 2.12 & 0.0005 & 8.67 & \textbf{27.24} & 0.20 & 9.35 & 6.7 & 24.07 & 0.61 & 6.47 & 1.64 & 6.77 & \textbf{0.12} & 8.41             \\ \midrule
% 			20 Newsgroups             & Global-Pool CNN                &  725.82 & 13.71 & \textbf{25.93} &  719.83 & 15.30 & 28.07 &  731.31 & 12.69 & 28.63 &  940.70 & \textbf{4.52} & 30.80 &  \textbf{487.82} & 16.55 & 27.88         \\ 
			\bottomrule
		\end{tabular}%
	}
	\caption{Top-label calibration measure ECE ($S_{95}$) (\% score) and $|S_{95}|$ (percentage of total number of test samples with predictive confidences $\ge 0.95$) in comparison with various competing methods. We use $M=10$ bins for ECE ($S_{95}$) calculations. We outperform all the baselines across various popular benchmark datasets, and architectures in terms of calibration. While we do not outperform all calibration methods in terms of $|S_{95}|$,it is to be noted that we obtain a higher $|S_{95}|$ than (\flsd, \mdca).}
	\label{tab:S95-all-methods}
%\vspace{-10mm}
\end{table*}

% Additionally, we divide the predicted confidences of the test set instances into $M=10$ bins, and plot out the number of instances falling into each bin as a histogram.

The results here are presented for the ResNet50 model,
we provide some additional results with the ResNet32 model in 
the supplementary material
that provides a tabular depiction of the results
shown here in \cref{tab:sce-all-methods} and \cref{tab:S95-all-methods}.
We observe a similar calibration performance for the ResNet32 model.

\smallskip
\mypara{Test Error} \cref{tab:sce-all-methods} also compares the Test Error (\te) obtained by the models trained using our proposed methods against all other \sota approaches. The metrics show that our proposed method achieves the best calibration performance without sacrificing on prediction accuracy (\te).

%\begin{wrapfigure}{r}{6cm}

% \smallskip
%\mypara{Reliability Diagrams} Please refer to the teaser image in
%\cref{fig:teaser} where we compare our proposed approach against models trained with Cross-Entropy, Label Smoothing (\ls), \dca \cite{dcapaper}, Focal Loss (\fl) \cite{ogfocalloss}, Brier Score (\bs) \cite{brierloss}, \flsd \cite{focallosspaper} as well as \mmce \cite{kumarpaper}.

% \smallskip
\mypara{Refinement} Refinement also measures trust in \dnns as the degree of separation between a network's correct and incorrect predictions.  \cref{tab:sce-all-methods} shows our proposed method is refined in addition to being calibrated.

% Buffer if there is space
%If $S^p = \{\mathbf{x} | \hat{y} = y; (\mathbf{x}, y_i) \in D_{te}\}$ denotes the set of all instances classified correctly, and similarly $S^n = \{\mathbf{x} | \hat{y} \neq y; (\mathbf{x}, y) \in D_{te}\}$ denotes the set of instances classified incorrectly where instance-label pairs $(\mathbf{x}, y)$ are samples from the test set $D_{te}$. The Neural networks predictions are considered refined if $c > c'$ $\forall \mathbf{x} \in S^p, \mathbf{x}' \in S^n$, where $c, c'$ are the predicted confidences of test set instances $\mathbf{x}, \mathbf{x}'$ respectively. 

\begin{figure*}
    \centering
        \includegraphics[width=\linewidth]{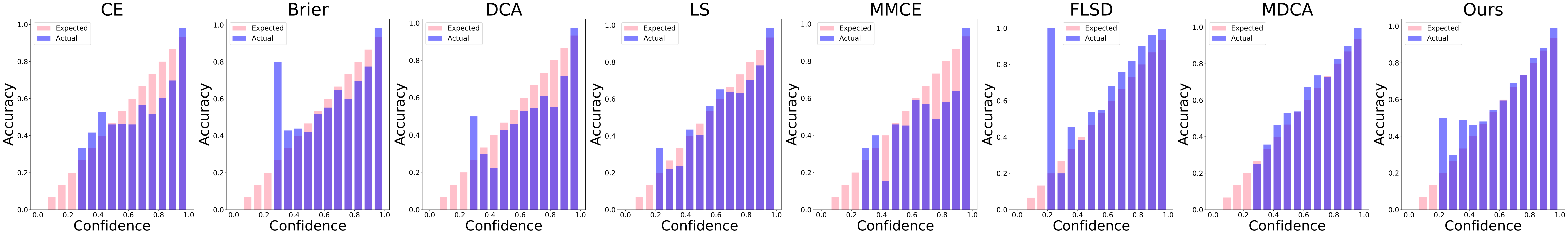}
    \caption{\textbf{Reliability diagrams our proposed calibration technique
    against the state-of-the-art methods}: A ResNet50 classifier is trained on CIFAR-10 using 
    \nll, Brier-score~\cite{brierloss}, DCA~\cite{dcapaper}), Label-smoothing~\cite{labelsmoothinghelp}, \mmce~\cite{kumarpaper}, \flsd~\cite{focallosspaper}, \mdca~\cite{StitchInTime} and our proposed method, respectively. 
    Reliability diagrams as a measure of calibration (When a \dnn is perfectly calibrated, the accuracy for each bin would be equal to that of the confidence of that particular bin, therefore all the bars would lie on $y=x$ line. If bars are below the $y=x$ line means \dnn is over-confident. If the bars are above the $y=x$ line, this means that the \dnn is considered under-confident). 
    Notice that \dnn trained using \nll, Brier-score, \dca, \ls, \mmce is over-confident 
    whereas a \dnn with \flsd is slightly under-confident. 
    \dnn trained using our method results in \textbf{calibrated and confident} predictions making our approach that uses dynamic train-time pruning appealing for practical deployment.}
    
    \label{fig:my_label}
\end{figure*}

\subsection{Ablation Study}\label{sec:ablations}
Our proposed approach has $3$ new additions over pre-existing \sota approaches. These are: (1) An auxiliary loss function, the Huber Loss that is targeted to reduce the top-label calibration; (2) A pruning step, based on predicted confidences; (3) Smoothing the confidences using an \ema to enhance calibration performance. We study the effect of these individual components in this section. %\cref{tab:prune_ema_ablations} \textcolor{red}{Rishabh fix this} reports the \ece, \ece$S_{95}$ and $|S_{95}|$ of these individual contributions.

\begin{figure}[!htb]
\centering
    \includegraphics[width=0.65\linewidth]{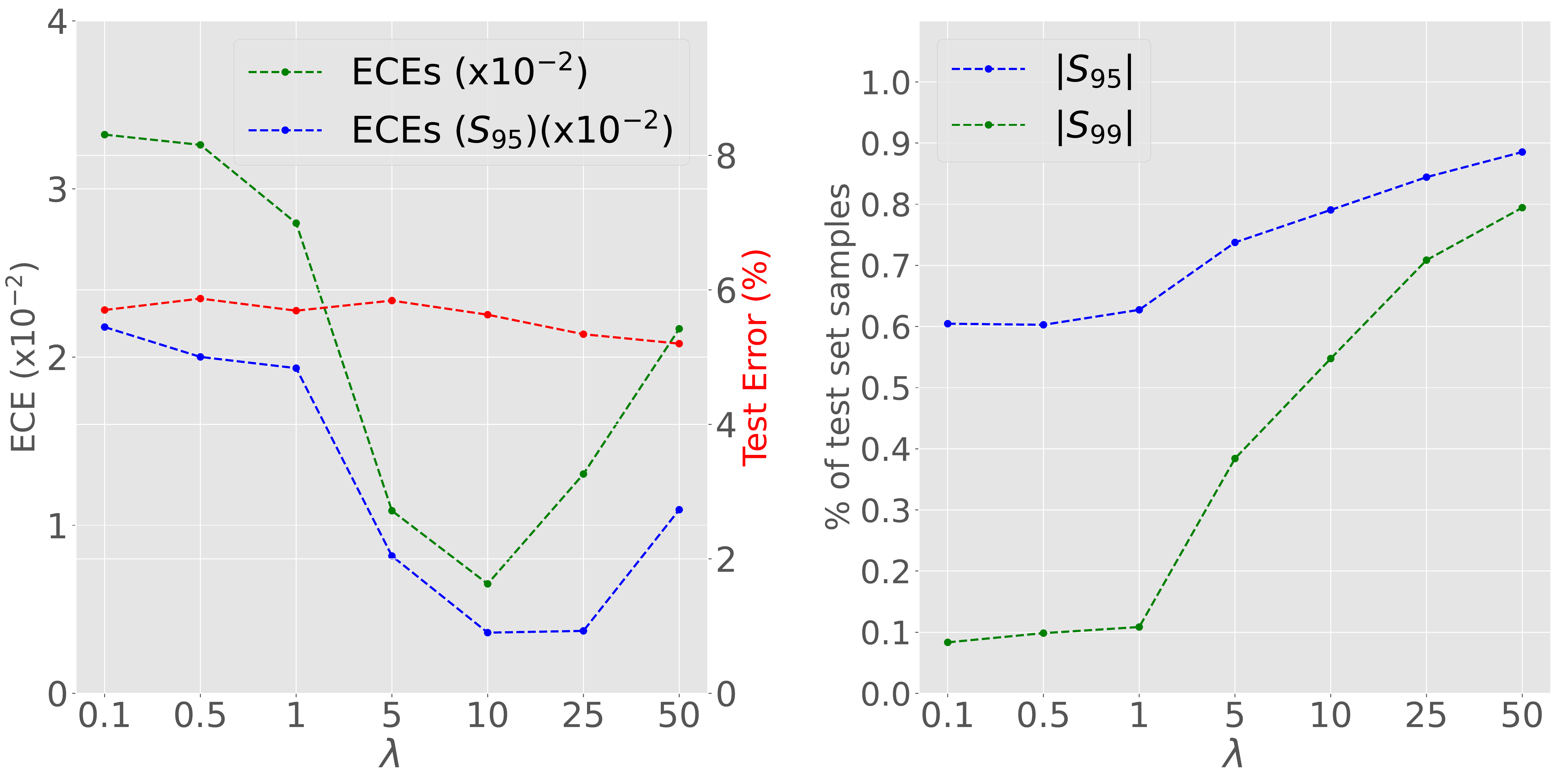}
    \caption{\textbf{Effect of varying the regularization factor, $\lambda$, on training ResNet50 model on CIFAR-10 dataset}. \textbf{left:} Plots out the \ece, \ece ($S_{95}$) and \te vs $\lambda$. \textbf{right:} Studying the effect of $|S_{95}|$ and $|S_{99}|$ on varying $\lambda$. Lowest \ece and \ece($S_{95}$) are achieved at $\lambda=10$. For higher values of $\lambda$($\lambda > 10$), we notice an increase in $|S_{95}|$ and $|S_{99}|$, but reduction in \te and calibration gets worse with $\uparrow$ in $\lambda$.}
    \label{fig:abl_lmbda}
\end{figure}

% \input{All_sections/tab_abl_prune_ema}
% \smallskip
%\subsubsection{Effect of varying the regularization factor ($\lambda$) on Calibration Error}
\mypara{Effect of varying the regularization factor ($\lambda$) on Calibration Error}
\label{subsec:VaryLambda}
We vary the effect of regularization offered by our proposed auxiliary loss function by varying $\lambda$ in \cref{eq:fL+h}. Plotting out the \ece, \ece$(S_{95})$, \te, and $|S_{95}|$ and $|S_{99}|$ in \cref{fig:abl_lmbda} for a ResNet50 model trained on CIFAR-10, varying the regularization factor $\lambda$ from $\{0.1, 0.5, 1, 5, 10, 25, 50\}$, we gain an interesting insight into the effect of this auxiliary loss. We notice that there is a steady rise in $|S_{95}|$ and $|S_{99}|$ as $\lambda$ is increased. However, we see a minimum value of \ece and \ece$(S_{95})$ at $\lambda = 10$. We hypothesise that an auxiliary loss between confidences and accuracies on top of calibration losses is a push-pull mechanism, the auxiliary loss tries to minimise the calibration errors by pushing the confidences up, whereas the calibration losses minimise overconfidence by pulling the confidences down. For our CIFAR-10 experiments, $\lambda=10$ proved to be the optimal balance achieving the lowest calibration errors.

\begin{figure}[!htb]
    \centering
    \includegraphics[width=0.65\linewidth]{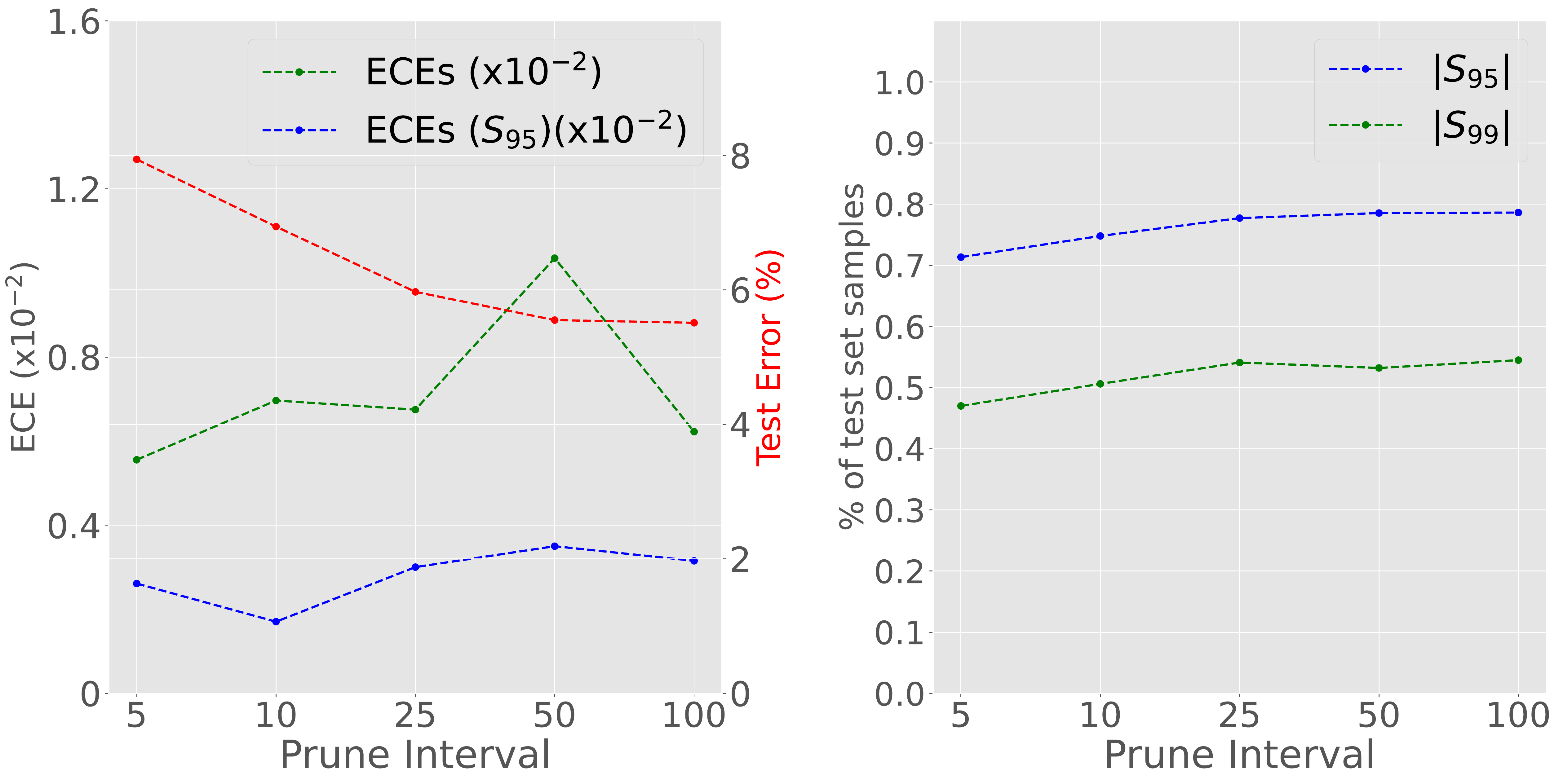}
    \caption{
    \textbf{Effect of varying the pruning interval on training} a ResNet50 on CIFAR-10 dataset. \textbf{Left:} Plots out the \ece, \ece($S_{95}$), and \te vs Pruning intervals. Pruning more frequently implies lower training time. \textbf{Right:} Studying the effect of $|S_{95}|$ and $|S_{99}|$ on varying the Prune Interval. Lowest \ece is achieved for pruning every $5$ epochs, and best \ece($S_{95}$) is achieved for pruning every $10$ epochs.}% Notice that the \te steadily declines with increase in pruning interval and $|S_{95}|$ and $|S_{99}|$ steadily increase.}
    \label{fig:abl_pruneFreq}
\end{figure}

% \begin{figure}[t]
% \includegraphics[width=\linewidth]{figures/abl_lmbda_cifar100.pdf}
% \caption{
%     Effect of varying the regularization factor ($\lambda$) on training a ResNet50 on CIFAR-100 dataset. (left) Plots out the \ece \ece($S95$) and \te vs $\lambda$. (right) Studying the effect of $|S95|$ and $|S99|$ on varying $\lambda$. Least \ece and \ece($S95$) are achieved at $\lambda=10$. Varying $\lambda$ doesnt have much affect on the test set error (\te). To note: Increasing $\lambda$ increases $|S95|$ and $|S99|$, but the calibration errors increase for $\lambda > 10$.
%     }
% \end{figure}
% \smallskip
\mypara{Effect of varying the pruning frequency on Calibration Error}
\label{subsec:pruneFreq}
Fig.~\ref{fig:abl_pruneFreq} shows how varying pruning frequency affects top label calibration. As the prune interval increases we see a monotonic decrease in the \te while least \ece is achieved when we prune every $5$ epochs. This hints the empirical evidence that increasing pruning frequency helps in reduced calibration error.

%\textcolor{red}{Is this consistent across other models? Rishabh: Yes, consistent across all experiments. Also, what fraction of samples are used to achieve the least \ece? Rishabh: I'll add a timing chart in a while.
% \smallskip
\mypara{Effect of using an \ema score to track evolution of prediction confidences}
\label{subsec:emaEvol}

Please refer to supplementary.

%\textcolor{red}{We need to write something here. Cross refernce to the relevant figure. }

%\textcolor{red}{write a paragraph with cross ref to figures}
% \begin{figure}[t]
% \includegraphics[width=\linewidth]{figures/abl_ema_cifar100.pdf}
% \caption{
%     Effect of varying the EMA factor ($\alpha$) on training a ResNet50 on CIFAR-100 dataset. (left) Plots out the \ece \ece($S95$) and \te vs $\alpha$. (right) Studying the effect of $|S95|$ and $|S99|$ on varying $\alpha$
%     }
% \end{figure}

\begin{figure}[!htb]
    \centering
    \includegraphics[width=0.65\linewidth]{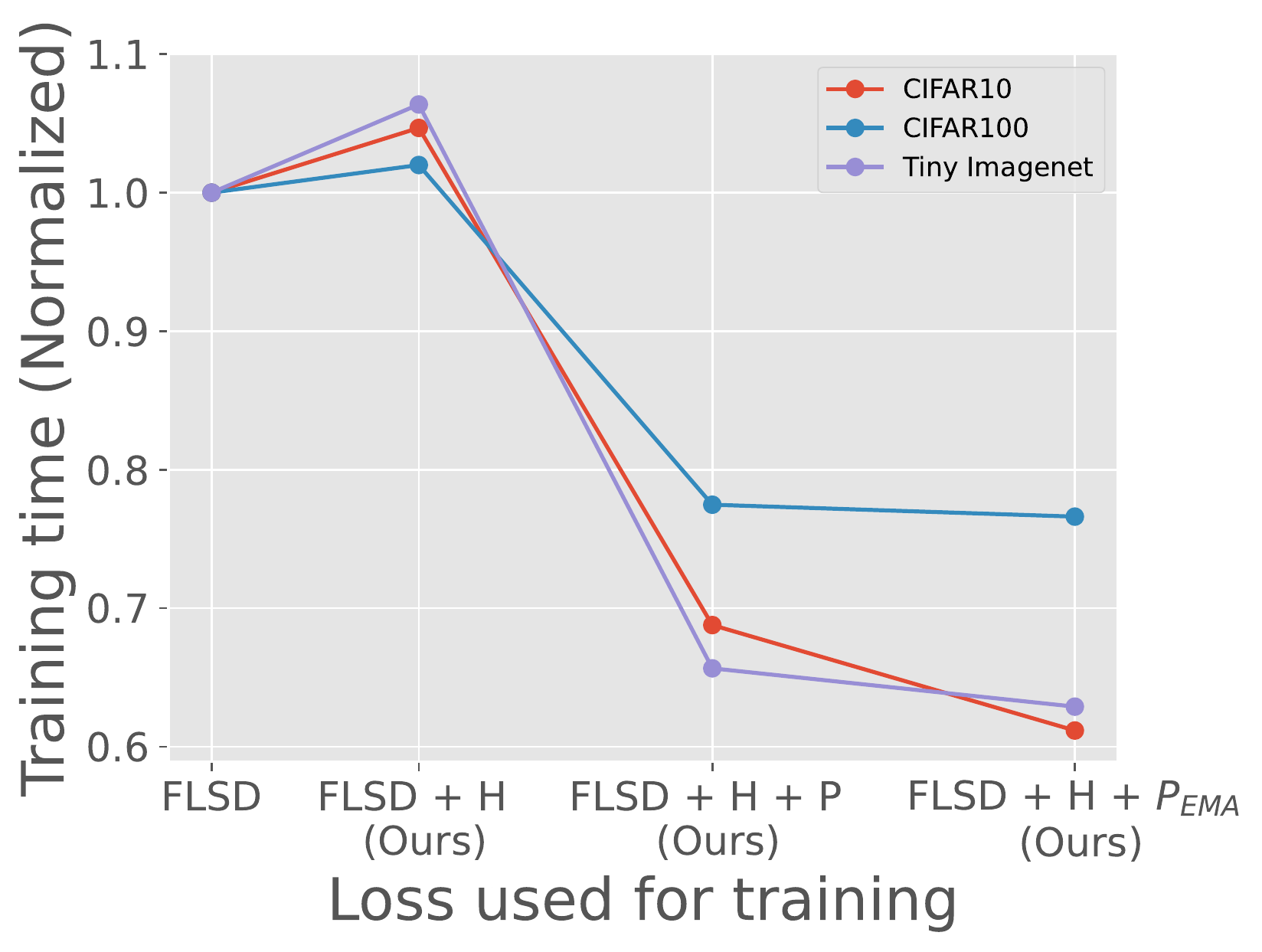}
    \label{fig:abl_ema}
    \caption{\textbf{Comparison of training times using our proposed pruning strategy}. FLSD~\cite{focallosspaper}, FLSD+H (Ours \cref{eq:fL+h}), FLSD+H+P (Ours: \cref{eq:fL+h} with Pruning)  and FLSD+H+$\text{P}_{\text{EMA}}$ (Ours: \cref{eq:fL+h} with Pruning with exponential moving average) for a ResNet50 trained on three different datasets. Upon using our pruning strategy, we see a reduction of ~20\% in training time with the CIFAR100 dataset. The train time reduction while using Tiny Imagenet and CIFAR10 is ~40\%. Lower training times of our proposed pruning strategies make our strategy appealing for a practical viewpoint.
    }
% \end{wrapfigure}
\end{figure}

% \smallskip
\mypara{Reduction in training time using our proposed pruning strategy}
\label{subsec:trainingtime}
When compared to FLSD \cite{focallosspaper}, the training time of proposed approach sees a reduction of $\textbf{40\%}$ in case of CIFAR10 and Tiny Imagenet Datasets, with a reduction of $\textbf{20\%}$ in case of CIFAR100. Recalibration for train time calibration methods implies retraining the model on the entire train dataset. Such recalibration efforts can be unappealing given the high training times and compute required to do so. We ensure our methods provide confident and calibrated models, while reducing training time, and hence the carbon footprint significantly.

% With a significant reduction in training time in addition to \sota calibration, we ensure that our methods provide confident and calibrated models with a reduction in the carbon footprint.

% \begin{wrapfigure}{r}{5cm}

%\vspace{-10mm}

\section{Conclusion}
\label{sec:conclusion}

We have introduced an efficient train-time calibration method without much trade off in accuracy of \dnns. We have made two contributions: first, we propose a differentiable loss term that can be used effectively in gradient descent optimisation used extensively in \dnn classifiers; second, our proposed dynamic data pruning strategy not only enhances legitimate high confidence samples to enhance trust in \dnn classifiers but also reduce the training time for calibration. We shed light on why pruning data facilitates high-confidence samples that aid in \dnn calibration. %In future we would like to extend this approach in pixel level classification and large medical imaging classification tasks, test its effectiveness for other non-visual domains.

% Suppl Material
%--------------
\clearpage

\appendix

\textbf{Supplementary Material: Calibrating Deep Neural Networks using Explicit Regularisation and Dynamic Data Pruning}

\author{Rishabh Patra$^{1}$\textsuperscript{\textsection} \quad  Ramya Hebbalaguppe$^{2}$\textsuperscript{\textsection} \quad  Tirtharaj Dash$^{1}$ \quad Gautam Shroff$^{2}$ \quad Lovekesh Vig$^{2}$\\
$^{1}$ APPCAIR, BITS Pilani, Goa Campus \quad
$^{2}$ TCS Research, New Delhi\\
%$^{*}$ First author\\
% {\tt\small f20180348@goa.bits-pilani.ac.in, ramya.hebbalaguppe@tcs.com}
% {\tt\small f20180348@goa.bits-pilani.ac.in}
% {\tt\small ramya.hebbalaguppe@tcs.com}
% {\tt\small tirtharaj@goa.bits-pilani.ac.in}
% {\tt\small gautam.shroff@tcs.com}
% {\tt\small lovekesh.vig@tcs.com}
}
\maketitle 
\begingroup\renewcommand\thefootnote{\textsection}
\footnotetext{Equal contribution}
\endgroup

\maketitle        

In this supplementary document,
We provide an detailed training algorithm corresponding to Procedure 2 in the main text. Further,
we provide some results of calibration for the ResNet32 model in addition to the results in the main text. Our datasets, codes and the 
resulting models, of all our experiments (shown in the main paper and in
the supplementary), will be made available publicly after acceptance.

\section{Datasets}
% \subsection{Datasets}
We validate our proposed approach on benchmark datasets for image classification. We chose CIFAR-10/100 datasets
\footnote{\url{https://www.cs.toronto.edu/~kriz/cifar.html}}, MendelyV2 (Medical image classification~\cite{kermany2018labeled}), SVHN\footnote{\url{http://ufldl.stanford.edu/housenumbers/}}, and Tiny ImageNet\footnote{\url{https://image-net.org/}}. 
In all our experiments, we calibrate ResNet-50 \cite{resnetpaper} and measure the 
calibration performances using our proposed calibration technique and several other existing techniques. For all experiments, the train set is split into $2$ mutually exclusive sets: (a) training set containing $90\%$ of samples and (b) validation set: $10\%$ of the samples. The same validation set is used for post-hoc calibration.

\section{A detailed pruning based learning procedure}
We provide additionally a detailed version of Procedure 2 from the main text. Procedure~\ref{proc:learn_elab} details each step of our pruning based learning procedure, as used to obtain calibration DNNs with reduced training times, translating to reduced carbon footprint and easier recalibration.

\begin{algorithm*}[!htb]
	\caption{Our pruning-based learning procedure. The procedure takes as inputs: 
	A dataset of $n$ instances: $D = \{(\mathbf{x}_i, y_i)\}_{i=1}^{n}$,
	An untrained neural network $\mathcal{N}$ with structure $\pi$ and parameters $\bm{\theta}$,
	Maximum number of training epochs: $MaxEpochs$,
	Batch-size: $b \leq n$, 
	Focal loss parameter: $\gamma$,
	Huber Loss parameter: $\alpha$,
	Regularization parameter: $\lambda$,
	Learning rate for SGD: $\eta$,
	Weight decay parameter for SGD: $\beta$,
	EMA factor for smoothing: $\kappa$,
	Prune fraction: $\epsilon \in (0,100)$, 
	A set of pruning epochs during training: $\mathbf{ep}$; 
	and returns: a trained model.
	The procedure assumes a parameter update procedure
	\text{\sc BackPropWithSGD}.}
	\label{proc:learn_elab}
	\begin{algorithmic}[1]
		\Procedure{TrainDNNwithDataPruning}{$D$,$\mathcal{N}$,$\pi$,$\bm{\theta}$,$MaxEpochs$,$b$,$\gamma$,$\alpha$,$\lambda$,$\eta$,$\beta$,$\kappa$,$\epsilon$,$\mathbf{ep}$}
		
		\State Let $De = \{(\mathbf{x}_i, y_i, 0)\}_{i=1}^{n}$ 
		where $(\mathbf{x}_i, y_i) \in D\}$, $i \in \{1,\dots,n\}$
 		\State Number of training batches: $nb=\left\lceil\frac{n}{b}\right\rceil$
 		\State Let $B_1,\dots,B_{nb}$ be the mini-batches of data instances from $De$
 		\State Initialise $\bm{\theta}$ to small random numbers
        \For{training epoch $ep$ in $\{1,\dots,MaxEpochs\}$}
            \For{$B_i \in \{B_1,\dots,B_{nb}\}$}
                \State Mean accuracy in batch $i$: ${acc} = 0$
                \State Mean confidence in batch $i$: ${conf} = 0$
                \State Mean focal loss: $\mathcal{L}_{FL} = 0$
            
                %obtain mean acuracy and mean confidence for this batch
                \For{each $(\mathbf{x}_k,y_k,e_k) \in B_i$}
                    \State $\hat{\bm{y}} = \mathcal{N}(\mathbf{x}_k;(\pi,\bm{\theta}))$
                    \State $\hat{y} = \arg \max_{i} \hat{\bm{y}}$
                    \State $c = \max(\hat{\bm{y}})$
                    \State ${acc} = {acc} + \mathbb{I}({\hat{y} = y_k})$
                    \State ${conf} = {conf} + c$
                    \State $\mathcal{L}_{FL} = \mathcal{L}_{FL} + \text{\sc ComputeFocalLoss}(\mathrm{onehot(y_k),\hat{\bm{y}}},\gamma)$
                    \State $e_k = \kappa c + (1-\kappa)e_k$)
                \EndFor
                
                %compute accuracy and confidence for this batch
                \State ${acc} = acc / b$
                \State ${conf} = conf / b$
                \State $\mathcal{L}_{FL} = \mathcal{L}_{FL} / b$
                
                %calculate losses
                \State Calculate Huber loss: $\mathcal{L}_{H} = \text{\sc ComputeHuberLoss}(acc,conf,\alpha)$
                \State Calculate total loss: $\mathcal{L}_{total} = \mathcal{L}_{FL} + \lambda\mathcal{L}_{H}$
                
                %update parameters of the network
                \State Update parameters of $\mathcal{N}$: $\bm{\theta} = \text{\sc BackPropWithSGD}(\mathcal{L}_{total}$,$\pi$,$\bm{\theta}$,$\eta$,$\beta$)
            \EndFor    
            \State Update the instances in $De$ with updated \ema-scores
            computed above
            \If{epoch $ep \in \mathbf{ep}$}
                \State $De = \text{\sc PruneUsingEMA}(De,\epsilon)$
            \EndIf
        \EndFor
		\EndProcedure
	\end{algorithmic}
\end{algorithm*}

\section{Additional Results with ResNet32}
\label{sec:tm}

\subsection{Experiments Parameters}

The experimental parameters remain largely the same as in our ResNet50 experiments.
For CIFAR10, we train the models for a total of $160$ epochs using an initial learning rate of $0.1$. The learning rate is reduced by a factor of $10$ at the $80^{th}$ and $120^{th}$ epochs. The \dnn was optimized using Stochastic Gradient Descent (SGD) with momentum $0.9$ and weight decay set at $0.0005$. Further, the images belonging to the trainset are augmented using random center cropping, and horizontal flips.
For CIFAR100, the models are trained for a total of $200$ epochs with a learning rate of $0.1$, reduced by a factor of $10$ at the $100^{th}$ and $150^{th}$ epochs. The batch size is set to 1024. Other parameters for training on CIFAR100 are the same ones used for CIFAR10.
For Tiny-Imagenet, we follow the same training procedure used by \cite{focallosspaper}. The models are trained for $100$ epochs, with a batch size of 1024.
For the Huber loss hyperparameter $\alpha$, 
we perform a grid search over values $\{0.001, 0.005, 0.01, 0.05, 0.1\}$. The setting $\alpha=0.005$ gave the best calibration results across all datasets, and hence we use the same value for all the experiments. 
% Other ablations regarding the Pruning Intervals, Regularization factor ($\lambda$), EMA factor ($\alpha$) can be found in section on Ablation Study .All experiments are performed on a single Nvidia-V100 GPU, with the exception of Tiny-imagenet experiments, for which 4 Nvidia-V100 GPUs were used to fit the larger batch-size into GPU memory.
For the regularization parameter $\lambda$, we perform a grid search over $\{0.1, 0.5, 1, 5, 10, 25, 50\}$, choosing $\lambda$ with the least \ece and \ece $(S_{95})$. In our ResNet32 experiments, we find that $\lambda=25$ gives the least calibration error in both the metrics for CIFAR10. For CIFAR100, $\lambda=5$ is the optimal parameter for low calibration errors.
To set the pruning frequency, we again perform a grid search over $\{5, 10, 25, 50, 100\}$. We find that pruning every $25$ epochs is optimal for CIFAR10, whereas pruning every 50 epochs is optimal for CIFAR100. Needless to say, we identify our optimal parameters as those which provide the least calibration error.

\mypara{Effect of using an \ema score to track evolution of prediction confidences}
\label{subsec:emaEvol}

\begin{figure}
\centering
\includegraphics[width=0.9\linewidth]{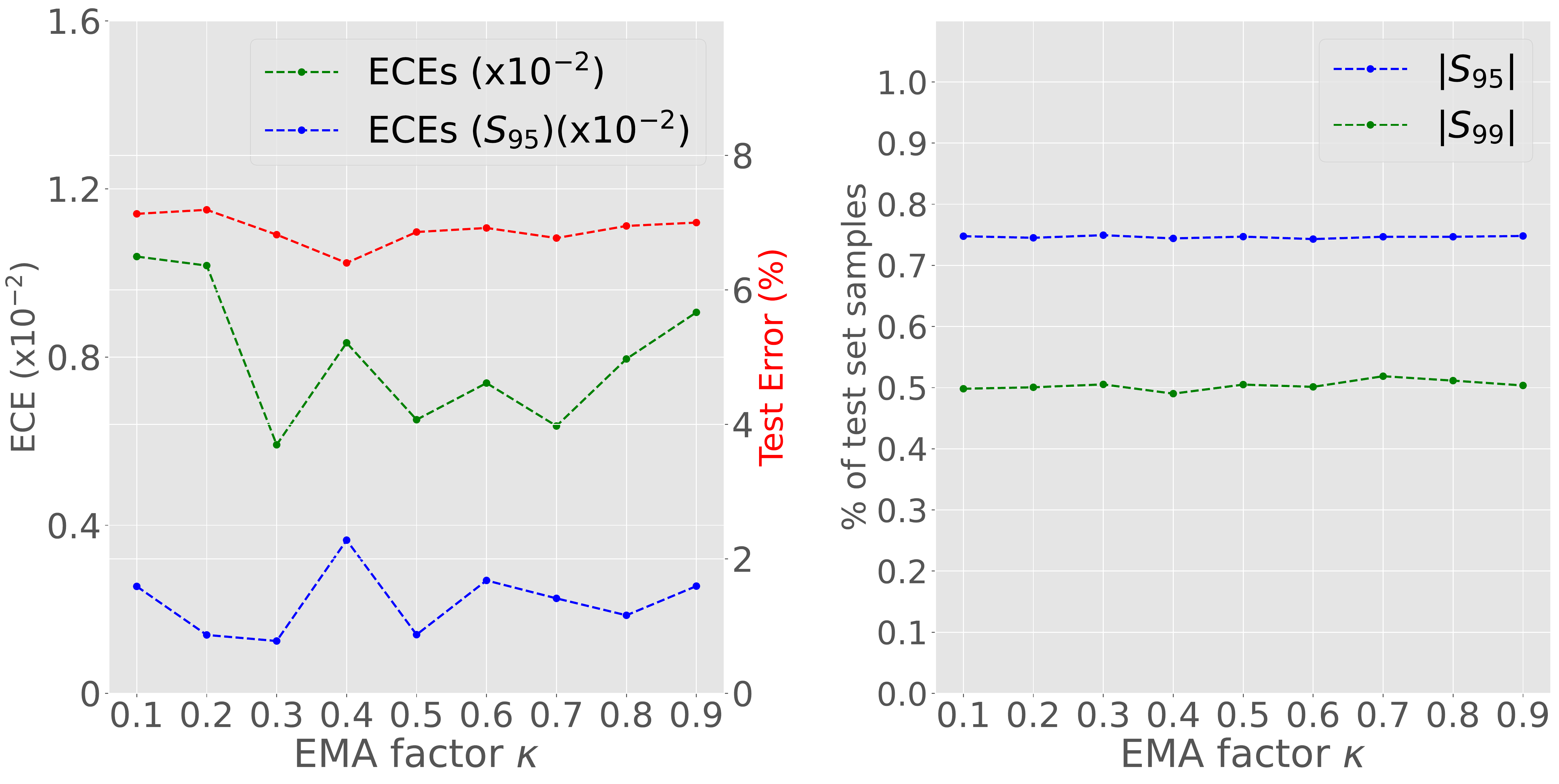}
\label{fig:abl_ema_1}
\caption{
    \textbf{Effect of varying the EMA factor} ($\kappa$) on training a ResNet50 on CIFAR-10 dataset. \textbf{Left:} Plots out the \ece, \ece($S_{95}$), and \te vs. \ema factor, $\kappa$. \textbf{Right:} Study of the effect of $|S_{95}|$ and $|S_{99}|$ on varying $\kappa$. The best \ece and \ece($|S_{95}|$) are achieved at $\kappa=0.3$. Varying $\kappa$ has no noticeable effect of $|S_{95}|$ and $|S_{99}|$} %\textcolor{red}{expand observation on left and right figures}.
    %}
\end{figure}
We show how the \te, \ece, \ece ($S_{95}$), $|S_{95}|$ and $S_{99}$ vary with increasing the \ema factor $\kappa$. There isn't any noticeable trend of metrics across varying $\kappa$. However, using an \ema score to track confidences across epochs, and subsequently using this \ema score to prune training instances is a practically appealing method since it produces near perfect calibration of the high confidence instances.

\subsection{Experimental Results}

\begin{table*}[!htb]
	\centering
	\resizebox{\linewidth}{!}{%
		\begin{tabular}{cc||ccc|ccc|ccc|ccc|ccc|ccc||ccc}
			\toprule
			&  &  \multicolumn{3}{c|}{\textbf{BS \cite{brierloss}}} & \multicolumn{3}{c|}{\textbf{DCA \cite{dcapaper}}}   & \multicolumn{3}{c|}{\textbf{LS \cite{originallabelsmoothing}}} &
			\multicolumn{3}{c|}{\textbf{MMCE \cite{kumarpaper}}} &
			\multicolumn{3}{c|}{\textbf{FLSD \cite{focallosspaper}}}  &
			\multicolumn{3}{c||}{\textbf{FL + MDCA \cite{StitchInTime}}} &
			\multicolumn{3}{c}{\textbf{Ours ({FLSD+H+$\text{P}_{\text{EMA}}$})}} \\
			\multirow{-2}{*}{\textbf{Dataset}} &
			\multirow{-2}{*}{\textbf{Model}} &

			ECE &
			TE &
			AUROC &
			
			ECE &
			TE &
			AUROC &
			
			ECE &
			TE &
			AUROC &
			
			ECE &
			TE &
			AUROC &
			
			ECE &
			TE &
			AUROC &
			
			ECE &
			TE &
			AUROC &
			
			ECE & 
			TE &    
			AUROC \\
			
			\hline
			\midrule
			CIFAR10  & ResNet32               & 2.83 & 7.67 & 90.35 & 3.58 & 17.13 & 87.39 & 3.06 & 7.71 & 85.25 & 4.16 & 7.36 & 90.81 & 5.27 & 7.82 & 91.24 & \textbf{0.93} & \textbf{7.18} & \textbf{91.62} & 1.10 & 8.33 & 91.42       \\ 
% 			& ResNet50                   & 2.11 & 5.69 & 91.76 & 3.45 & 5.09 & 93.03 & 3.43 & 4.80 & 82.71 & 3.12 & \textbf{5.05} & 93.55 & 3.37 & 5.61 & 93.39 & 1.06 & 5.32 & \textbf{93.60} & \textbf{0.59} & 6.82 & 91.64       \\ \midrule
			CIFAR100 & ResNet32               & 7.87 & 36.78 & \textbf{85.67} & 7.82 & 42.91 & 82.46 & 7.44 & 34.66 & 83.52 & 15.09 & \textbf{33.33} & 84.48 & 3.28 & 35.7 & 82.91 & \textbf{3.06} & 35.47 & 83.39 & 3.18 & 35.41 & 82.36       \\
% 			& ResNet50                & 5.18 & 26.74 & 86.12 & 6.7 & 45.43 & 86.75 & 3.72 & \textbf{25.67} & 86.45 & 10.52 & 25.68 & \textbf{86.81} & 3.25 & 26.56 & 85.88 & 2.96 & 26.51 & 86.32 & \textbf{1.74} & 26.57 & 86.17   \\ \midrule
% 			\multirow{2}{*}{SVHN}     & ResNet20               &   &  &  &   &  &  &   &  &  &   &  &  &   &  &  &  &  &  &       \\
% 			SVHN & ResNet50                  & 1.22 & 3.36 & 88.50 & 0.71 & 3.83 & 92.04 & 4.25 & \textbf{3.25} & 84.57 & 1.86 & 3.46 & 87.75 & 12.58 & 3.61 & 88.38 & 7.96 & 3.74 & 89.84 & \textbf{0.38} & 3.99 & \textbf{90.79}     \\ \midrule
% 			Mendeley V2               & ResNet50                & 22.99 & 27.04 & 64.91 & 21.42 & 22.59 & 56.60 & 15.81 & 23.39 & 66.19 & 21.73 & 25.48 & 72.19 & 12.59 & 31.41 & 71.99 & 17.09 & 27.08 & 73.48 & \textbf{11.69} & \textbf{22.27} & \textbf{75.96}   \\ \midrule
% %			Kather5000                & ResNet34                &  32.76 & 8.64 & 16.50 &  27.13 & 4.49 & 16.80 &  47.35 & 16.63 & 17.90 &  57.87 & 22.03 & 19.30 &  41.39 & 13.45 & 16.60              \\ \midrule
% 			Tiny-ImageNet             & ResNet50                & 0.36 & 99.32 & 78.37 & 13.98 & 47.59 & \textbf{82.99} & 15.55 & \textbf{45.28} & 82.80 & 9.33 & 46.29 & 82.20 & 4.42 & 46.76 & 82.75 & 4.67 & 46.14 & 82.50 & \textbf{1.44} & 51.26 & 82.00         \\ \midrule
% 			20 Newsgroups             & Global-Pool CNN                &  725.82 & 13.71 & \textbf{25.93} &  719.83 & 15.30 & 28.07 &  731.31 & 12.69 & 28.63 &  940.70 & \textbf{4.52} & 30.80 &  \textbf{487.82} & 16.55 & 27.88         \\ 
			\bottomrule
		\end{tabular}%
	}
	\caption{Calibration measure \ece (\%) score), Test Error (\te) (\%) and AUROC (refinement) in comparison with various competing methods. We use $M=10$ bins for \ece calculation. We outperform most of the baselines across various popular benchmark datasets, and architectures in terms of calibration, while maintaining a similar accuracy and a similiar refinement (AUROC.)}
	\label{tab:sce_32-all-methods}
%\vspace{-6mm}
\end{table*}

\begin{table*}[!htb]
	\centering
	\resizebox{\linewidth}{!}{%
		\begin{tabular}{cc|cc|cc|cc|cc|cc|cc||cc}
			\toprule
			&  &  \multicolumn{2}{c|}{\textbf{BS \cite{brierloss}}} & 
			\multicolumn{2}{c|}{\textbf{DCA \cite{dcapaper}}} & 
			\multicolumn{2}{c|}{\textbf{LS \cite{hinton2015distilling}}} &
			\multicolumn{2}{c|}{\textbf{MMCE \cite{kumarpaper}}} &
			\multicolumn{2}{c|}{\textbf{FLSD \cite{focallosspaper}}}  &
			\multicolumn{2}{c||}{\textbf{FL + MDCA \cite{StitchInTime}}} &
			\multicolumn{2}{c}{\textbf{Ours ({FLSD+H+$\text{P}_{\text{EMA}}$})}} \\
			\multirow{-2}{*}{\textbf{Dataset}} &
			\multirow{-2}{*}{\textbf{Model}} &

			ECE ($S_{95}$) &
			$|S_{95}|$ &
			
			ECE ($S_{95}$) &
			$|S_{95}|$ &
			
			ECE ($S_{95}$) &
			$|S_{95}|$ &
			
			ECE ($S_{95}$) &
			$|S_{95}|$ &
			
			ECE ($S_{95}$) &
			$|S_{95}|$ &
			
			ECE ($S_{95}$) &
			$|S_{95}|$ &
			
			ECE ($S_{95}$) &
			$|S_{95}|$  \\
			\hline
			\midrule
			CIFAR10 & ResNet32                & 1.49 & 82.44 & 1.29 & 54.74 & 1.55 & 74.46 & 2.54 & \textbf{87.80} & 2.36 & 40.52 & 1.56 & 57.63 & \textbf{0.002} & 72.36    \\
			 %& ResNet50               & 0.87 & 87.05 & 2.27 & \textbf{93.30}  & 3.01 & 84.82 & 1.89 & 92.83 & 2.06 & 60.27 & 0.80 & 76.71 & \textbf{0.12} & 74.94          \\ \midrule
			CIFAR100 & ResNet32                & 3.47 & 33.34 & 2.26 & 20.28 & 3.31 & 28.40 & 7.84 & \textbf{46.63} & 0.54 & 16.47 & 0.11 & 17.07 & \textbf{0.008} & 17.37          \\
			 %& ResNet50               & 2.92 & 45.24 & 1.5 & 15.13 & 1.29 & 28.7 & 5.44 & \textbf{58.96} & 1.22 & 20.12 & 1.35 & 22.23 & \textbf{0.01} & 33.1      \\ \midrule
% 			\multirow{2}{*}{SVHN}     & ResNet20               &   &  &  &   &  &  &   &  &  &   &  &  &  &         \\
% 			SVHN & ResNet50              & 0.68 & 92.02 & 0.36 & 88.5 & 3.72 & 48.32 & 1.14 & \textbf{94.64} & 3.55 & 0.41 & 3.75 & 13.23 & \textbf{0.045} & 86.39        \\ \midrule
% 			Mendeley V2               & ResNet50               & 21.95 & 74.84 & 22.04 & \textbf{93.10}  & 13.22 & 58.49 & 21.11 & 82.05 & 7.92 & 11.58 & 7.70 & 53.52 & \textbf{1.94} & 22.27    \\ \midrule
%			Kather5000                & ResNet34                &  32.76 & 8.64 & 16.50 &  27.13 & 4.49 & 16.80 &  47.35 & 16.63 & 17.90 &  57.87 & 22.03 & 19.30 &  41.39 & 13.45 & 16.60              \\ \midrule
% 			Tiny-ImageNet             & ResNet50                & 2.12 & 0.0005 & 8.67 & \textbf{27.24} & 0.20 & 9.35 & 6.7 & 24.07 & 0.61 & 6.47 & 1.64 & 6.77 & \textbf{0.12} & 8.41             \\ \midrule
% 			20 Newsgroups             & Global-Pool CNN                &  725.82 & 13.71 & \textbf{25.93} &  719.83 & 15.30 & 28.07 &  731.31 & 12.69 & 28.63 &  940.70 & \textbf{4.52} & 30.80 &  \textbf{487.82} & 16.55 & 27.88         \\ 
			\bottomrule
		\end{tabular}%
	}
	\caption{Top-label calibration measure ECE ($S_{95}$) (\% score) and $|S_{95}|$ (percentage of total number of test samples with predictive confidences $\ge 0.95$) in comparison with various competing methods. We use $M=10$ bins for ECE ($S_{95}$) calculations. We outperform all the baselines across various popular benchmark datasets, and architectures in terms of calibration. While we do not outperform all calibration methods in terms of $|S_{95}|$,it is to be noted that we obtain a higher $|S_{95}|$ than (\flsd, \mdca).}
	\label{tab:S95_32-all-methods}
%\vspace{-10mm}
\end{table*}

\cref{tab:sce_32-all-methods} shows the \te, \ece and AUROC for refinement of the ResNet32 models trained on CIFAR10. We notice that \mdca outperforms our proposed approach in terms of \ece. However, it is to be noted that our proposed approach is better calibrated than all the other \sota approaches. \cref{tab:S95_32-all-methods} shows the \ece ($S_{95}$) and $|S_{95}|$ for our ResNet32 experiments. Here, we obtain the least \ece ($S_{95}$) against all other \sota approaches, which is in alignment with our claims of our approach being appealing for practical scenarios. Further, we obtain better $|S_{95}|$ than other Focal loss based methods (ie. \mdca, \flsd).

{\small
\bibliographystyle{ieee_fullname}
\bibliography{citations}

\begin{thebibliography}{10}\itemsep=-1pt

\bibitem{bohdal2021meta}
Ondrej Bohdal, Yongxin Yang, and Timothy Hospedales.
\newblock Meta-calibration: Meta-learning of model calibration using
  differentiable expected calibration error.
\newblock In {\em ICML Uncertainity in Deep Learning Workshop}, 2021.

\bibitem{brierloss}
Glenn~W Brier et~al.
\newblock Verification of forecasts expressed in terms of probability.
\newblock {\em Monthly weather review}, 78(1):1--3, 1950.

\bibitem{rkhskernel}
Arthur Gretton.
\newblock Introduction to rkhs, and some simple kernel algorithms.
\newblock {\em Adv. Top. Mach. Learn. Lecture Conducted from University College
  London}, 16:5--3, 2013.

\bibitem{guo2017calibration}
Chuan Guo, Geoff Pleiss, Yu Sun, and Kilian~Q Weinberger.
\newblock On calibration of modern neural networks.
\newblock In {\em ICML}, pages 1321--1330. PMLR, 2017.

\bibitem{resnetpaper}
Kaiming He, Xiangyu Zhang, Shaoqing Ren, and Jian Sun.
\newblock Deep residual learning for image recognition, 2015.

\bibitem{StitchInTime}
Ramya Hebbalaguppe, Jatin Prakash, Neelabh Madan, and Chetan Arora.
\newblock A stitch in time saves nine: A train-time regularizing loss for
  improved neural network calibration.
\newblock In {\em IEEE/CVF CVPR}, June 2022.

\bibitem{hinton2015distilling}
Geoffrey Hinton, Oriol Vinyals, and Jeff Dean.
\newblock Distilling the knowledge in a neural network.
\newblock {\em arXiv preprint arXiv:1503.02531}, 2015.

\bibitem{huber1992robust}
Peter~J Huber.
\newblock Robust estimation of a location parameter.
\newblock In {\em Breakthroughs in statistics}, pages 492--518. Springer, 1992.

\bibitem{kermany2018labeled}
Daniel Kermany, Kang Zhang, Michael Goldbaum, et~al.
\newblock Labeled optical coherence tomography (oct) and chest x-ray images for
  classification.
\newblock {\em Mendeley data}, 2(2), 2018.

\bibitem{Dirichlet}
Meelis Kull, Miquel Perello-Nieto, Markus K{\"a}ngsepp, Hao Song, Peter Flach,
  et~al.
\newblock Beyond temperature scaling: Obtaining well-calibrated multiclass
  probabilities with dirichlet calibration.
\newblock {\em arXiv preprint arXiv:1910.12656}, 2019.

\bibitem{beta-cal-paper}
Meelis Kull, Telmo Silva~Filho, and Peter Flach.
\newblock Beta calibration: a well-founded and easily implemented improvement
  on logistic calibration for binary classifiers.
\newblock In {\em Artificial Intelligence and Statistics}, pages 623--631.
  PMLR, 2017.

\bibitem{kumarpaper}
Aviral Kumar, Sunita Sarawagi, and Ujjwal Jain.
\newblock Trainable calibration measures for neural networks from kernel mean
  embeddings.
\newblock In {\em ICML}, pages 2805--2814, 2018.

\bibitem{dcapaper}
Gongbo Liang, Yu Zhang, Xiaoqin Wang, and Nathan Jacobs.
\newblock Improved trainable calibration method for neural networks on medical
  imaging classification.
\newblock {\em CoRR}, abs/2009.04057, 2020.

\bibitem{ogfocalloss}
Tsung-Yi Lin, Priya Goyal, Ross Girshick, Kaiming He, and Piotr Doll{\'a}r.
\newblock Focal loss for dense object detection.
\newblock In {\em Proceedings of the IEEE ICCV}, pages 2980--2988, 2017.

\bibitem{mahajan2020coviddiagnosis}
Kushagra Mahajan, Monika Sharma, Lovekesh Vig, Rishab Khincha, Soundarya
  Krishnan, Adithya Niranjan, Tirtharaj Dash, Ashwin Srinivasan, and Gautam
  Shroff.
\newblock {CovidDiagnosis: Deep Diagnosis of COVID-19 Patients Using Chest
  X-Rays}.
\newblock In {\em International Workshop on Thoracic Image Analysis}, pages
  61--73. Springer, 2020.

\bibitem{mukhoti2020calibrating}
Jishnu Mukhoti, Viveka Kulharia, Amartya Sanyal, Stuart Golodetz, Philip~HS
  Torr, and Puneet~K Dokania.
\newblock Calibrating deep neural networks using focal loss.
\newblock {\em arXiv preprint arXiv:2002.09437}, 2020.

\bibitem{focallosspaper}
Jishnu Mukhoti, Viveka Kulharia, Amartya Sanyal, Stuart Golodetz, Philip H.~S.
  Torr, and Puneet~K. Dokania.
\newblock Calibrating deep neural networks using focal loss, 2020.

\bibitem{labelsmoothinghelp}
Rafael M{\"u}ller, Simon Kornblith, and Geoffrey Hinton.
\newblock When does label smoothing help?
\newblock {\em arXiv preprint arXiv:1906.02629}, 2019.

\bibitem{can-u-trust}
Yaniv Ovadia, Emily Fertig, Jie Ren, Zachary Nado, David Sculley, Sebastian
  Nowozin, Joshua~V Dillon, Balaji Lakshminarayanan, and Jasper Snoek.
\newblock Can you trust your model's uncertainty? evaluating predictive
  uncertainty under dataset shift.
\newblock {\em arXiv preprint arXiv:1906.02530}, 2019.

\bibitem{datadietpaper}
Mansheej Paul, Surya Ganguli, and Gintare~Karolina Dziugaite.
\newblock Deep learning on a data diet: Finding important examples early in
  training.
\newblock {\em Advances in Neural Information Processing Systems}, 34, 2021.

\bibitem{pereyra2017regularizing}
Gabriel Pereyra, George Tucker, Jan Chorowski, {\L}ukasz Kaiser, and Geoffrey
  Hinton.
\newblock Regularizing neural networks by penalizing confident output
  distributions.
\newblock {\em arXiv preprint arXiv:1701.06548}, 2017.

\bibitem{platt1999probabilistic}
John Platt et~al.
\newblock Probabilistic outputs for support vector machines and comparisons to
  regularized likelihood methods.
\newblock {\em Advances in large margin classifiers}, 10(3):61--74, 1999.

\bibitem{singh2021deep}
Aditya Singh, Alessandro Bay, Biswa Sengupta, and Andrea Mirabile.
\newblock On deep neural network calibration by regularization and its impact
  on refinement.
\newblock {\em arXiv preprint arXiv:2106.09385}, 2021.

\bibitem{originallabelsmoothing}
Christian Szegedy, Vincent Vanhoucke, Sergey Ioffe, Jonathon Shlens, and
  Zbigniew Wojna.
\newblock Rethinking the inception architecture for computer vision, 2015.

\bibitem{forgettingpaper}
Mariya Toneva, Alessandro Sordoni, Remi~Tachet des Combes, Adam Trischler,
  Yoshua Bengio, and Geoffrey~J. Gordon.
\newblock An empirical study of example forgetting during deep neural network
  learning.
\newblock In {\em ICLR}, 2019.

\end{thebibliography}
}

\end{document}